\documentclass[10pt,final,journal,twocolumn,singlespaced]{IEEEtran}
\usepackage{algorithm}
\usepackage{algpseudocode}
\usepackage{amsmath}
\usepackage{amssymb}
\usepackage{bm}
\usepackage{booktabs}
\usepackage{color}
\usepackage{cite}
\usepackage{epsfig}
\usepackage{epstopdf}
\usepackage{flafter}
\usepackage{float}
\usepackage{ifthen}
\usepackage{multirow}
\usepackage{amsfonts}
\usepackage{makecell} 
\usepackage{pifont}
\usepackage{subfigure}
\usepackage{times}
\usepackage{threeparttable}
\usepackage{url} 
\usepackage{subfigure,graphicx}
\usepackage{float}
\usepackage{color,multirow}
\usepackage{makecell}
\usepackage{setspace}
\usepackage{cite} 
\usepackage{url} 
\usepackage{ifthen} 
\usepackage[T1]{fontenc}
\usepackage{graphicx}
\usepackage{changepage}
\usepackage[colorlinks,linkcolor=red,anchorcolor=gray,citecolor=green,urlcolor=blue]{hyperref}
\usepackage{graphics}
\usepackage{booktabs}

\begin{document}
\title{Unsupervised Hyperspectral Mixed Noise Removal Via Spatial-Spectral Constrained Deep Image Prior

\thanks{This research is supported by NSFC (No. 61876203, 61772003), the Applied Basic 
Research Project of Sichuan Province (No. 2021YJ0107), the Key Project of Applied Basic Research in Sichuan Province (No. 2020YJ0216), and National Key Research and Development Program of China (No. 2020YFA0714001).}
\thanks{Y.-S. Luo, X.-L. Zhao, and Y.-B. Zheng are with the Research Center for Image and Vision Computing, School of Mathematical Sciences, University of Electronic Science and Technology of China, Chengdu, P.R.China (e-mail: yisiluo1221@foxmail.com; xlzhao122003@163.com; zhengyubang@163.com).}\thanks{T.-X. Jiang is with the FinTech Innovation Center, the Financial Intelligence and Financial Engineering Research Key Laboratory of Sichuan Province, School of Economic Information Engineering, Southwestern University of Finance and Economics, Chengdu, P.R.China (e-mail: taixiangjiang@gmail.com).}
\thanks{Y. Chang is with the Artificial Intelligence Research Center, Peng Cheng Laboratory, Shenzhen, P.R.China (e-mail: yichang@hust.edu.cn).}}
\author{Yi-Si Luo,
Xi-Le Zhao, \IEEEmembership{Member, IEEE},
Tai-Xiang Jiang, \IEEEmembership{Member, IEEE},
Yu-Bang Zheng, \IEEEmembership{Student Member, IEEE},
and Yi Chang, \IEEEmembership{Member, IEEE}}
\maketitle
\begin{abstract}
Recently, convolutional neural network (CNN)-based methods are proposed for hyperspectral images (HSIs) denoising. Among them, unsupervised methods such as the deep image prior (DIP) have received much attention because these methods do not require any training data. However, DIP suffers from the semi-convergence behavior, i.e., the iteration of DIP needs to terminate by referring to the ground-truth image at the optimal iteration point. In this paper, we propose the spatial-spectral constrained deep image prior (S2DIP) for HSI mixed noise removal. Specifically, we incorporate DIP with a spatial-spectral total variation (SSTV) term to fully preserve the spatial-spectral local smoothness of the HSI and an $\ell_1$-norm term to capture the complex sparse noise. The proposed S2DIP jointly leverages the expressive power brought from the deep CNN without any training data and exploits the HSI and noise structures via hand-crafted priors. Thus, our method avoids the semi-convergence behavior, showing higher stabilities than DIP. Meanwhile, our method largely enhances the HSI denoising ability of DIP. To tackle the proposed denoising model, we develop an alternating direction multiplier method algorithm. Extensive experiments demonstrate that the proposed S2DIP outperforms optimization-based and supervised CNN-based state-of-the-art HSI denoising methods. 
\end{abstract}
\IEEEpeerreviewmaketitle
\section{Introduction} \label{sec:Int}
\begin{figure}[t]
\begin{center}
\setlength{\abovecaptionskip}{0.cm}
\includegraphics [width = 1\linewidth,height = 6.3cm]{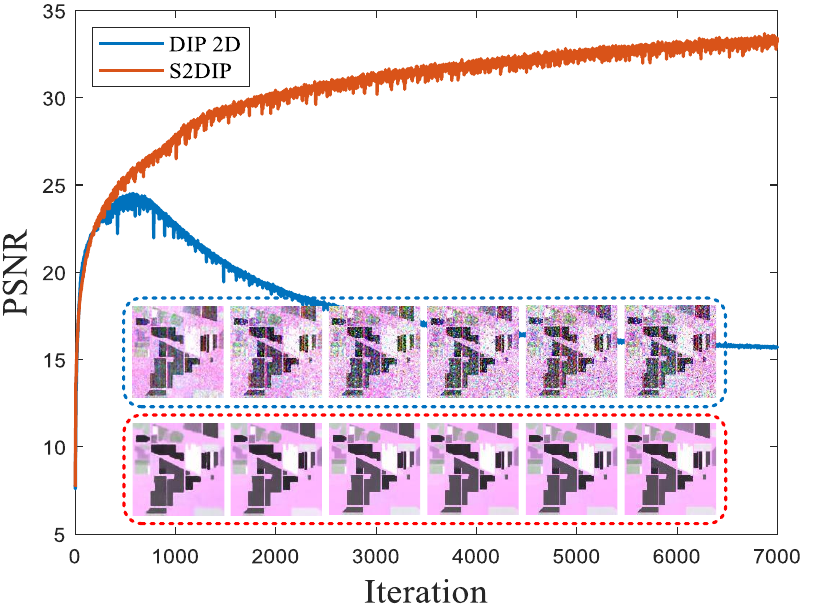}
\vspace{-0.6cm}
\end{center}
\caption{The history of PSNR values of the denoising results by DIP 2D \cite{DeepHS} and S2DIP with respect to the iterations. Compared to DIP 2D, S2DIP is more stable and achieves better results for HSI mixed noise removal. \label{semi_con}}
\end{figure}
\IEEEPARstart{H}{yperspectral} images (HSIs) contain abundant spatial and spectral information and thus can be utilized into various applications, such as object detection \cite{Wang2016,Liu2017Target}, classification
\cite{Jia2018,TNNLS_class1,Meng_2}, and so on. Due to the imaging environment, HSIs are inevitably corrupted by mixed noise. The complex noise seriously affects subsequent applications as well as visual qualities. 
Thus, HSI denoising is considered an essential technique in hyperspectral imaging.\par 
Earlier HSI denoising methods are dominated by model-based techniques \cite{Wang_2,He_2,Meng_3,Wang_1}, such as total variation (TV)-based methods\cite{ASSTV,SSTV,HSSTV,DIP-tv}, sparsity-based methods \cite{sparse1,sparse2}, non-local-based methods \cite{BM4D}, dictionary learning-based methods \cite{Dic_HSI,CVPR_Dic}, and matrix/tensor low-rankness-based methods \cite{LRMR,tvLRMR,LRTDTV,ChangYi2017,Low_rank_JSTSP,XieQi_PAMI,TNNLS_Meng}. These methods exploit the intrinsic structure of HSIs via optimization-based models to deal with the HSI denoising. With meticulously selected hand-crafted priors, recent model-based methods \cite{LRMR,LRTDTV,LLRGTV} have achieved state-of-the-art performance. These state-of-the-art methods mainly consider the matrix/tensor low-rankness and form the HSI denoising \cite{Lina_1,Lina_2} as a low-rank matrix/tensor recovery problem. Other hand-crafted priors such as TV \cite{He_3,He_4} can be combined in such low-rank model and algorithms like alternating direction multiplier method (ADMM) can be utilized to address the denoising model.\par 
Inspired by the success of deep learning in inverse problems in imaging \cite{TIP_2017,NIPS2016,ResNet,Deep_HSI_sharpening}, convolutional neural network (CNN)-based methods \cite{TNNLS_2014,2019Dong,HSI-DeNet,HSID-CNN,TowardUni,TNNLS_2020_HSI} have emerged for HSI denoising in last few years. Generally, most of the CNN-based methods learn the noisy-to-denoising mapping driven by abundant training data \cite{HSID-CNN,2019Dong}, which leads to great performance for HSI denoising on specific datasets. Nevertheless, the effectiveness of supervised CNN-based methods critically depends on the diversity and quantity of training data. However, hyperspectral data is limited and the real noise of HSIs is complex. Thus, it is difficult to guarantee high-quality denoising results under the realist complex noise scenario of HSIs, where the underlying assumption is not held in the training data. \par 
Recently, an unsupervised image restoration method deep image prior (DIP) \cite{DIP} was proposed. The DIP employs the CNN and optimizes its learnable parameters by targeting the observed image as the network output with a randomly given input, demonstrating that the CNN itself can represent a well-reconstructed image by the iterative process without any training data. Afterward, Sidorov et al. extended the DIP into HSI restoration \cite{DeepHS}, which yields decent performance on HSI denoising, inpainting, and super-resolution. \par 
However, DIP suffers from the semi-convergence behavior, which refers to the behavior of an iterative method that the PSNR value begins to increase at early iterations and, after a certain ``optimal'' iteration, the PSNR value begins to decrease, see Fig. \ref{semi_con}. Hence, the iteration needs to terminate by referring to the ground-truth image at the point of the highest PSNR value, before fitting the noise. Meanwhile, DIP has relatively weak performance for HSI denoising compared with existing state-of-the-art methods \cite{HSID-CNN,LRTDTV,LRMR}. We attribute these deficiencies of DIP to two aspects. First, DIP lacks specific characterization on HSI intrinsic structures. HSIs have unique spatial-spectral structures, which require specific prior terms to characterize. Second, DIP lacks robust noise modeling. In real applications, the noise of HSI is very complex and various. Only considering Gaussian noise and neglect other types of noise is not enough to completely remove the complex noise in HSI. 
\par 
In this paper, we propose the spatial-spectral constrained deep image prior (S2DIP) for HSI mixed noise removal. We combine DIP with prior information on both HSI and hyperspectral noise. Specifically, we employ a spatial-spectral total variation (SSTV) term to explore the spatial-spectral local smoothness of HSIs and an $\ell_1$-norm term to capture the sparse noise for mixed noise removal. To tackle the proposed denoising model, we develop an ADMM algorithm. Meanwhile, we elaborately design an automatic stopping criterion to verify the stability of S2DIP. The DIP, HSI prior, and noise prior are complementary to each other and are organically combined to benefit each other. The proposed S2DIP could favorably address the semi-convergence behavior of DIP and largely improves the effectiveness and robustness for HSI mixed noise removal. \par
Compared with traditional optimization-based HSI denoising methods, the proposed S2DIP capitalizes the expressive power from the CNN and holds higher model representation ability. Compared with supervised CNN-based methods, the proposed S2DIP does not need training data and has a better generalization ability for diverse HSI data with various complex noise. \par
We summarize the contributions of this paper as:
\begin{itemize}
	\item We propose the S2DIP for HSI mixed noise removal.  Specifically, we incorporate DIP with an SSTV term to fully explore the spatial-spectral local smooth prior of the underlying HSI and an $\ell_1$-norm sparse term to capture the sparse noise. Our method avoids the semi-convergence behavior and is more stable than DIP. Meanwhile, with the comprehensive consideration of the deep prior, HSI prior, and sparse noise prior, the proposed S2DIP could achieve promising HSI denoising results.
\item To tackle the proposed denoising model, we develop an efficient ADMM algorithm. In company, we elaborately design an automatic stopping criterion without referring to the ground-truth image.
\item Extensive experiments on HSIs, MSIs, and videos validate the effectiveness and generalization ability of the proposed method. The proposed method outperforms optimization-based and CNN-based state-of-the-art methods. Moreover, HSI classification results after the denoising are reported to validate the effectiveness of the proposed method.
\end{itemize}\par
The rest of this paper is organized as follows. In Sec. \ref{sec:Rel}, we introduce some related work. In Sec. \ref{sec:Pro}, we introduce the proposed S2DIP. In Sec. \ref{sec:Exp}, we carry out the experimental results. Sec. \ref{sec:Dis} provides some discussions. Finally, Sec. \ref{sec:Con} concludes this paper.
\begin{figure*}[!t]
\centering
\setlength{\abovecaptionskip}{0.1cm}
\includegraphics [width = 1\linewidth]{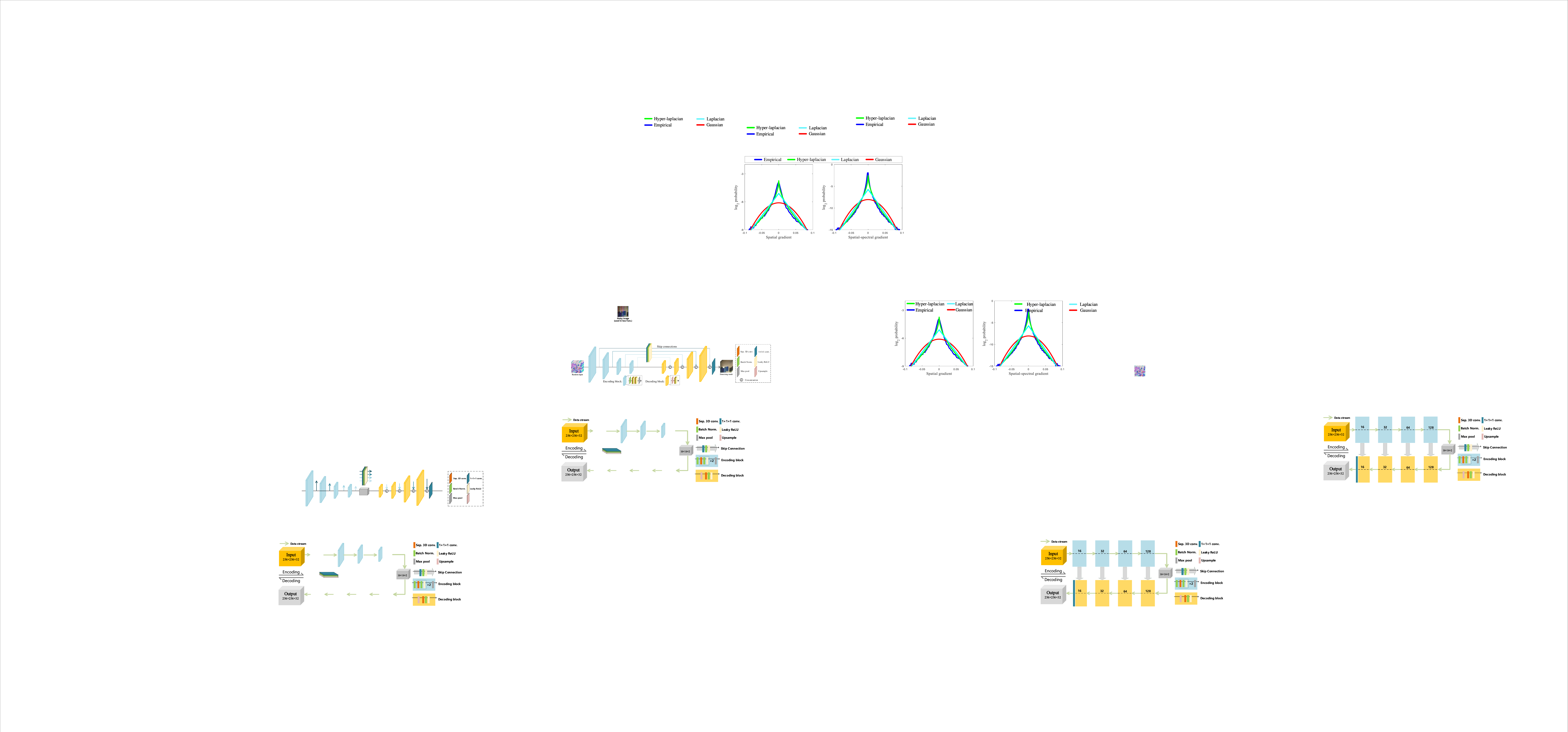}
\vspace{-0.3cm}
\caption{The network structure used in this work. The separable 3D convolution-driven encoder-decoder network aims at preferably representing the spatial and spectral features under the unsupervised condition.
\label{net}}\vspace{-0.1cm}
\end{figure*}
\section{Related Work} \label{sec:Rel}
\subsection{Model-Based Methods for HSI Denoising}
Traditional HSI denoising methods are dominated by model-based methods. These methods, including but not limited to TV-based methods \cite{ASSTV,SSTV,HSSTV,DIP-tv,Yuan_2,He_1,8894531}, sparsity-based methods \cite{sparse1,sparse2,Lina_2}, and matrix/tensor low-rankness-based methods \cite{LRMR,tvLRMR,LRTDTV,ChangYi2017,Low_rank_JSTSP,XieQi_PAMI,TNNLS_Meng,Lina_1,Zhang_1}, consider the HSI prior information and establish optimization model and corresponding algorithms for denoising. For instance, Zhang {et al.} \cite{LRMR} vectorized each band of the HSI as a column and unfolds them as a matrix, and then considered the low-rank property on the unfolded matrix. Wang {et al.}\cite{LRTDTV} utilized the tensor decomposition, which delivered the global tensor low-rankness. The TV regularization was considered along with the tensor decomposition. Such methods achieve state-of-the-art performance due to the comprehensive consideration of the HSI prior information. \par
\subsection{CNN-Based Methods for HSI Denoising} 
In recent years, the CNN-based approaches for HSI denoising \cite{2019Dong,HSI-DeNet,HSID-CNN,TowardUni,TNNLS_2020_HSI,Yuan_1,Meng_1} have emerged and present state-of-the-art performance. For instance, Yuan {et al.}\cite{HSID-CNN} employed the deep residual CNN to conduct HSI denoising. Dong {et al.}\cite{2019Dong} utilized the deep 3D encoder-decoder network for HSI denoising. Cao et al. \cite{Meng_1} used the deep spatial-spectral global reasoning network for HSI denoising. The core concept of these methods is to train a CNN with abundant pairs of training data $\{{\mathcal Y}, {\mathcal X}\}$, where $\mathcal Y$ denotes the dataset of noisy HSIs and $\mathcal X$ denotes the dataset of corresponding clean HSIs. The training process could be described as
\begin{equation}
\min_{\Theta}\mathcal{L}(f_{\Theta}(\mathcal{Y}),\mathcal{X}),
\end{equation} 
where $f_{\Theta}(\cdot)$ denotes a CNN with learnable parameters ${\Theta}$, and $\bf{\mathcal{L}}$ is the loss function. The well-trained CNN can be viewed as a noisy-to-denoising mapping. Due to the powerful nonlinear modeling ability of the deep CNN, supervised methods have achieved promising results on specific datasets and specific noisy types.\par
\subsection{Deep Image Prior}
Recently, Ulyanov et al. \cite{DIP} proposed an unsupervised deep learning-based image restoration technique, named deep image prior. The DIP uses CNN to conduct image restoration without any training process. By targeting the degraded image as the network output with randomly generated network input, the CNN could remove the Gaussian noise from the degraded image with appropriate iteration steps. Afterward, Sidorov et al. \cite{DeepHS} extended DIP into HSI restoration. \par 
Similar to the supervised methods, the optimization process of DIP for HSI \cite{DeepHS} restoration is formulated as
\begin{equation}
\min_{\Theta}{\frac{1}{N}}\lVert f_{\Theta}({\mathcal Z})-{\mathcal Y}\rVert_{\ell_2}^2,\label{dip_eq}
\end{equation} 
where ${\mathcal Z}\in {\mathbb R}^{H\times W\times B}$ denotes the randomly generated network input and $\mathcal Y$ denotes the noisy HSI. The mean square error (MSE) is adopted as the loss function, where $N=H\times W\times B$ denotes the number of total pixels. The gradient descent algorithm is adopted to iteratively optimize the CNN. Obviously, the optimization process will finally converge to a noisy HSI $f_{{\Theta}}({\mathcal Z})$, which is almost structurally identical with the observation ${\mathcal Y}$. However, it has been discovered that the network will firstly fit the signal part of the observation, and then fit the noisy part \cite{DIP}. Thus, the denoising of ${\mathcal Y}$ can be achieved by stopping the iteration at appropriate steps before the network fitting noise, and the denoising result is obtained by ${\mathcal X} = f_{\Theta^*}({\mathcal Z})$ with $\Theta^*$ denotes the network parameters after appropriate iterations. This is also a key limitation of DIP, where the denoising iterative process must be stopped by referring to the ground-truth image at the optimal point, see Fig. \ref{semi_con} for an example of DIP for HSI denoising. We refer to this limitation as semi-convergence. In this paper, we avoid this issue by combining DIP with spatial-spectral priors of HSI and the sparse prior of hyperspectral noise.
\section{The Proposed S2DIP} \label{sec:Pro}
This section introduces the proposed S2DIP. First, we introduce the degradation model of hyperspectral mixed noise. Then, we give the network structure of the proposed method and form the optimization model for HSI denoising. To address the proposed model, We develop an ADMM algorithm. Finally, we propose an automatic stopping criterion to be used for verifying the stability of S2DIP, which addresses the semi-convergence of DIP.
\subsection{Degradation Model}
The noise in HSI is complex and various. Only considering Gaussian noise \cite{DeepHS} is hard to comprehensively remove the complex noise. This motivates us to improve the generalization ability of DIP by considering more robust noise modeling. In this work, we consider the noisy HSI as an addition of the clean HSI, the Gaussian noise, and the sparse structural noise \cite{LRTDTV,HSSTV}. The degradation process can be formulated as
\begin{equation}
{\mathcal Y}={\mathcal X}+{\mathcal N}+{\mathcal S},
\end{equation}
where ${\mathcal Y}\in {\mathbb R}^{H\times W\times B}$ denotes the observed noisy HSI, ${\mathcal X}\in{\mathbb R}^{H\times W\times B}$ denotes the underlying clean HSI, ${\mathcal N}\in{\mathbb R}^{H\times W\times B}$ denotes the Gaussian noise, and ${\mathcal S}\in{\mathbb R}^{H\times W\times B}$ denotes the sparse structural noise. Our method will use the deep prior of CNN to generate ${\mathcal X}$ and fully consider the spatial-spectral prior information of ${\mathcal X}$ and the sparsity of ${\mathcal S}$ for mixed noise removal.
\subsection{Network Architecture of S2DIP}
We employ a U-Net with skip connections for HSI denoising \cite{DIP,DeepHS}, see Fig. \ref{net}. The input data is the random noise, which goes through the encoder and progressively outcomes embedded three-dimensional features. The embedded features go through the decoder to reconstruct the clean HSI. \par
In \cite{DeepHS}, the author tried both 2D and 3D convolution in the U-Net for HSI denoising. However, it appears that the performance of 3D convolution is inferior to that of 2D convolution. In this paper, we use the separable 3D convolution \cite{P3D}, which is shown to be more capable to encode the HSI than 3D convolution \cite{2019Dong}. The separable 3D convolution represents the spatial and spectral information using 2D and 1D kernels respectively so that better representation of the HSI can be obtained, see more details in \cite{2019Dong}. Here, we employ four layers in both encoding and decoding stages. We denote the proposed separable 3D U-Net by $f_\Theta(\cdot)$ with $\Theta$ refers to the learnable parameters. We randomly initialize the parameters of the network. The network parameters are iteratively and unsupervisedly updated (see Sec. \ref{Sec:Alg} and Algorithm \ref{alg}). Except for the observed HSI, no extra training data is needed.\par
\subsection{Optimization Model of S2DIP}\label{opt} 
{\bf Spatial-spectral constraint}:
The optimization model of DIP (\ref{dip_eq}) aims at fitting the noisy HSI while the noise could be removed using the ``deep prior'' of CNN. However, it is unavoidable for DIP that the CNN will eventually fit the noise, which causes semi-convergence. To address this issue, we propose to use the SSTV regularization to fully preserve the spatial-spectral local smoothness of the network output so that spatial-spectral consistency can be ensured. The SSTV term is expected to address the semi-convergence and simultaneously improve the denoising performance. \par
Formally, for a three-way tensor ${\mathcal X}\in{\mathbb R}^{H\times W\times B}$, its difference tensors ${\nabla_x}{\mathcal X}\in{\mathbb R}^{(H-1)\times W\times B}$, ${\nabla_y}{\mathcal X}\in{\mathbb R}^{H\times (W-1)\times B}$, and ${\nabla_z}{\mathcal X}\in{\mathbb R}^{H\times W\times (B-1)}$ are defined as
\begin{equation}
\left\{
\begin{aligned}
&{\nabla_x}{\mathcal X}(i,j,k)={\mathcal X}(i+1,j,k)-{\mathcal X}(i,j,k)\\
&{\nabla_y}{\mathcal X}(i,j,k)={\mathcal X}(i,j+1,k)-{\mathcal X}(i,j,k)\\
&{\nabla_z}{\mathcal X}(i,j,k)={\mathcal X}(i,j,k+1)-{\mathcal X}(i,j,k),
\end{aligned}
\right.
\end{equation}
where ${\mathcal X}(i,j,k)$ denotes the $(i,j,k)$-th element of $\mathcal X$, and $\nabla_x$, $\nabla_y$, and $\nabla_z$ denote the finite difference operators on the vertical direction, horizontal direction, and spectral direction, respectively. The TV of ${\mathcal X}$ is given by
\begin{equation}
\lVert{\mathcal X}\rVert_{\rm TV} = \lVert{\nabla_x}{\mathcal X}\rVert_{\ell_1}+\lVert{\nabla_y}{\mathcal X}\rVert_{\ell_1}.
\label{2DTV}
\end{equation}
Further considering the spatial-spectral local smoothness, the SSTV \cite{SSTV} of $\mathcal X$ is given by
\begin{equation}
\lVert{\mathcal X}\rVert_{\rm SSTV} = \lVert{\nabla_x}({\nabla_z}{\mathcal X})\rVert_{\ell_1}+\lVert{\nabla_y}({\nabla_z}{\mathcal X})\rVert_{\ell_1}.
\label{SSTV}
\end{equation}\par
The TV considers the spatial local smoothness and SSTV considers the spatial-spectral local smoothness. To faithfully utilize both of their properties, we introduce both TV and SSTV in the optimization model of DIP to explore the spatial and spatial-spectral local smooth prior of HSIs under unsupervised conditions. \par
{\bf Sparse noise modeling}:
The DIP model (\ref{dip_eq}) only considers the Gaussian noise. In real applications, the noise of HSI is complex and various. In this paper, we further consider the sparse noise to improve the generalization ability of DIP for more robust denoising. The sparse noise contains impulse noise and stripes that are commonly existing in HSIs. We consider minimizing the $\ell_1$-norm of $\mathcal S$ to enforce the sparsity on ${\mathcal S}$ so that the sparse noise can be faithfully separated from the clean HSI. \par
Based on the analysis of spatial-spectral constraints and sparse noise modeling, the proposed optimization model for HSI mixed noise removal is formulated as
\begin{equation}
\begin{split}
\min_{\Theta,{\mathcal S}}&\lVert {\mathcal Y}-{\mathcal X}-{\mathcal S}\rVert_{F}^2+\alpha_1\lVert{\mathcal X}\rVert_{\rm TV}+\alpha_2\lVert{\mathcal X}\rVert_{\rm SSTV}\\&+\alpha_3\lVert{\mathcal S}\rVert_{\ell_1}\\{\rm s.t.\;} &{\mathcal X} = f_{\Theta}({\mathcal Z}),\label{main_model}
\end{split}
\end{equation} 
where $\alpha_i(i=1,2,3)$ are trade-off parameters. Here, $\lVert{\mathcal X}\rVert_{\rm TV}$ and $\lVert{\mathcal X}\rVert_{\rm SSTV}$ are the TV and SSTV regularizations, respectively. $\lVert{\mathcal S}\rVert_{\ell_1}$ is the $\ell_1$-norm of the sparse noise. $\lVert {\mathcal Y}-{\mathcal X}-{\mathcal S}\rVert_{F}^2$ is the fidelity term. The CNN $f_\Theta(\cdot)$ with random input $\mathcal Z$ is used to model the clean HSI ${\mathcal X}$.\par
In our model, the deep prior, HSI prior, and sparse noise prior are complementary to each other and are organically combined to remove the mixed noise in HSI.
\subsection{Algorithm}\label{Sec:Alg}
\begin{algorithm}[t]
\begin{spacing}{1.00}
\renewcommand\arraystretch{1.2}
\caption[Caption for LOF]{HSI Mixed Noise Removal Using S2DIP}\label{alg}
\begin{algorithmic}[1]
\renewcommand{\algorithmicrequire}{\textbf{Input:}} 
\Require
Noisy HSI ${\mathcal{Y}}$, $t_{max}=7000$, $r=0.001$;
\renewcommand{\algorithmicrequire}{\textbf{Initialization:}} 
\Require Randomly initialize $\Theta$, ${\Lambda_i}={\bf 0}$, $t=0$, $r'=r$;
\While {$t\leq t_{max}$ and $r'\geq r$}
\State Update $\mathbf{\mathcal{V}}_i$ $(i = 1,2,3,4)$ via (\ref{V});
\State Update $\mathbf{\mathcal{S}}$ via (\ref{eq_S});
\State Update ${\Theta}$ via (\ref{theta});
\State Update ${{\Lambda}}_i$ $(i = 1,2,3,4)$ via (\ref{D});
\State $r'=\frac{\lVert f_{\Theta^{t+1}}({\mathcal Z})-f_{\Theta^{t}}({\mathcal Z})\rVert_{F}^2}{\lVert f_{\Theta^{t}}({\mathcal Z})\rVert_{F}^2}$;
\State $t = t+1$;
\EndWhile
\renewcommand{\algorithmicrequire}{\textbf{Output:}}
\Require The denoising HSI ${\mathcal X} = f_{\Theta}(\mathcal{Z})$;
\end{algorithmic}
\end{spacing}
\end{algorithm}
To tackle the proposed model (\ref{main_model}), we develop an efficient ADMM algorithm. By introducing four auxiliary variables ${\mathcal V}_i(i=1,2,3,4)$, we can formulate (\ref{main_model}) as
\begin{equation}
    \begin{aligned}
\min_{\Theta,\mathcal{S}}&\left\lVert \mathcal{Y}-f_{\Theta}(\mathcal{Z})-\mathcal{S}
    \right\rVert_F^2+\alpha_1\left\lVert \mathcal{V}_1\right\rVert_{\ell_1}+\alpha_1\left\lVert \mathcal{V}_2\right\rVert_{\ell_1}\\&+\alpha_2\left\lVert \mathcal{V}_3\right\rVert_{\ell_1}+\alpha_2\left\lVert \mathcal{V}_4\right\rVert_{\ell_1}+\alpha_3\lVert{\mathcal S}\rVert_{\ell_1}\\
    \text{s.t.}\;\;\;&
\mathcal{V}_1=\nabla_{x}{(f_\Theta({\mathcal Z}))},
\;\mathcal{V}_2=\nabla_{y}{(f_\Theta({\mathcal Z}))},\\&
\;\mathcal{V}_3=\nabla_{x}\nabla_{z}{(f_\Theta({\mathcal Z}))},
\;\mathcal{V}_4=\nabla_{y}\nabla_{z}{(f_\Theta({\mathcal Z}))}.
\label{model_2}
    \end{aligned}
\end{equation} 
The augmented Lagrangian function of (\ref{model_2}) is
\begin{equation}
\begin{split}
&\mathcal{L}_{\mu}(\Theta,\mathcal{S},\mathcal{V}_i,\Lambda_i)\\&=
\lVert \mathcal{Y}-f_{\Theta}(\mathcal{Z})-\mathcal{S}\rVert_F^2
+\alpha_1\lVert \mathcal{V}_1\rVert_{\ell_1}
+\alpha_1\lVert \mathcal{V}_2\rVert_{\ell_1}
\\&+\alpha_2\lVert \mathcal{V}_3\rVert_{\ell_1}
+\alpha_2\lVert \mathcal{V}_4\rVert_{\ell_1}
+\frac{\mu}{2}\lVert\nabla_{x}{(f_\Theta({\mathcal Z}))}-\mathcal{V}_1\Vert_F^2\\&
+\frac{\mu}{2}\lVert\nabla_{y}{(f_\Theta({\mathcal Z}))}-\mathcal{V}_2\Vert_F^2
+\frac{\mu}{2}\lVert\nabla_x\nabla_{z}{(f_\Theta({\mathcal Z}))}-\mathcal{V}_3\Vert_F^2
\\&
+\frac{\mu}{2}\lVert\nabla_y\nabla_{z}{(f_\Theta({\mathcal Z}))}-\mathcal{V}_4\Vert_F^2
+\alpha_3\lVert{\mathcal S}\rVert_{\ell_1}\\&
+<\Lambda_1,\nabla_{x}{(f_\Theta({\mathcal Z}))}-\mathcal{V}_1>
+<\Lambda_2,\nabla_{y}{(f_\Theta({\mathcal Z}))}-\mathcal{V}_2>\\&
+<\Lambda_3,\nabla_{x}\nabla_{z}{(f_\Theta({\mathcal Z}))}-\mathcal{V}_3>\\&
+<\Lambda_4,\nabla_{y}\nabla_{z}{(f_\Theta({\mathcal Z}))}-\mathcal{V}_4>,\\
\end{split}
\end{equation}
where $\mu$ is the penalty parameter and $\Lambda_i(i=1,2,3,4)$ are the multipliers. Using the ADMM, the problem can be divided into the following sub-problems.\par
{\bf $\mathcal{V}$ Sub-problems}:
{The $\mathcal{V}_i(i=1,2,3,4)$ sub-problems are}
\begin{equation}
	\left\{
	\begin{aligned}
		&\min_{\mathcal{V}_1} \frac{\mu}{2}\left\lVert \nabla_{x}{(f_{\Theta^{t}}({\mathcal Z}))}+\frac{\Lambda_1^{t}}{\mu}-\mathcal{V}_1\right\Vert_F^2+\alpha_1\left\lVert \mathcal{V}_1\right\Vert_{\ell_1}\\
		&\min_{\mathcal{V}_2} \frac{\mu}{2}\left\lVert \nabla_{y}{(f_{\Theta^{t}}({\mathcal Z}))}+\frac{\Lambda_2^{t}}{\mu}-\mathcal{V}_2\right\Vert_F^2+\alpha_1\left\lVert \mathcal{V}_2\right\Vert_{\ell_1}\\
		&\min_{\mathcal{V}_3} \frac{\mu}{2}\left\lVert \nabla_{x}\nabla_{z}{(f_{\Theta^{t}}({\mathcal Z}))}+\frac{\Lambda_3^{t}}{\mu}-\mathcal{V}_3\right\Vert_F^2+\alpha_2\left\lVert \mathcal{V}_3\right\Vert_{\ell_1}\\
		&\min_{\mathcal{V}_4} \frac{\mu}{2}\left\lVert \nabla_{y}\nabla_{z}{(f_{\Theta^{t}}({\mathcal Z}))}+\frac{\Lambda_4^{t}}{\mu}-\mathcal{V}_4\right\Vert_F^2+\alpha_2\left\lVert \mathcal{V}_4\right\Vert_{\ell_1},
		\label{V}
	\end{aligned}
	\right.
\end{equation}
which can be exactly solved by
\begin{equation}
	\left\{
	\begin{aligned}
		&\mathcal{V}_1^{t+1}={\it Soft}_{\frac{\alpha_1}{\mu}}(\nabla_{x}{(f_{\Theta^{t}}({\mathcal Z}))}+\frac{\Lambda_1^{t}}{\mu})\\
		&\mathcal{V}_2^{t+1}={\it Soft}_{\frac{\alpha_1}{\mu}}(\nabla_{y}{(f_{\Theta^{t}}({\mathcal Z}))}+\frac{\Lambda_2^{t}}{\mu})\\
		&\mathcal{V}_3^{t+1}={\it Soft}_{\frac{\alpha_2}{\mu}}(\nabla_{x}\nabla_{z}{(f_{\Theta^{t}}({\mathcal Z}))}+\frac{\Lambda_3^{t}}{\mu})\\
		&\mathcal{V}_4^{t+1}={\it Soft}_{\frac{\alpha_2}{\mu}}(\nabla_{y}\nabla_{z}{(f_{\Theta^{t}}({\mathcal Z}))}+\frac{\Lambda_4^{t}}{\mu}),
		\label{V}
	\end{aligned}
	\right.
\end{equation}
where ${\it Soft}_v(\cdot)$ is the soft-thresholding operator defined as
\begin{equation}
	{\it Soft}_{v}(\mathcal{X})(i,j,k)={\rm sign}({\mathcal X}(i,j,k))(\max\{|{\mathcal X}(i,j,k)|-v, 0\}).
\end{equation}\par
{\bf $\mathcal{S}$ Sub-problem}:
The ${\mathcal S}$ sub-problem is
\begin{equation}
\min_{S}\lVert{\mathcal Y}-f_{\Theta^{t}}({\mathcal Z})-{\mathcal S}\rVert_F^2+\alpha_3\lVert\mathcal{S}\rVert_{\ell_1},
\end{equation}
which can be exactly solved by
\begin{equation}
\mathcal{S}^{t+1}={\it Soft}_{{2\alpha_3}}({\mathcal Y}-f_{\Theta^{t}}({\mathcal Z})).\label{eq_S}
\end{equation}\par
{\bf ${\Theta}$ Sub-problem}
The ${\Theta}$ sub-problem is
\begin{equation}
    \begin{aligned}
&\min_{\Theta}\left\lVert \mathcal{Y}-f_{\Theta}(\mathcal{Z})-\mathcal{S}^t
    \right\rVert_F^2
+\frac{\mu}{2}\big{(}\left\lVert\nabla_{x}(f_{\Theta}(\mathcal{Z}))-{\mathcal D}^t_1\right\Vert_F^2\\&
+\left\lVert\nabla_{y}(f_{\Theta}(\mathcal{Z}))-{\mathcal D}^t_2\right\Vert_F^2+\left\lVert\nabla_{x}\nabla_{z}(f_{\Theta}(\mathcal{Z}))-{\mathcal D}^t_3\right\Vert_F^2\\&+\left\lVert\nabla_{y}\nabla_{z}(f_{\Theta}(\mathcal{Z}))-{\mathcal D}^t_4\right\Vert_F^2\big{)},
\label{theta}
    \end{aligned}
\end{equation}
where ${\mathcal D}^t_i = \mathcal{V}_i^t-\frac{\Lambda_i^t}{\mu}$. We adopt the adaptive moment estimation (Adam) \cite{ADAM} to update $\Theta$. Specifically, we employ one gradient descent step by using the Adam in each update. The loss functions are all $F$-norm-based functions, so the gradient can be easily computed in current deep learning frameworks. \par
{\bf $\Lambda$ Updating}
{The Lagrange multipliers are updated as
\begin{equation}
\left\{
    \begin{aligned}
&{\Lambda}_1^{t+1}={\Lambda}_1^t+\mu(\nabla_{x}(f_{\Theta^t}(\mathcal{Z}))-\mathcal{V}_1^t)\\
&{\Lambda}_2^{t+1}={\Lambda}_2^t+\mu(\nabla_{y}(f_{\Theta^t}(\mathcal{Z}))-\mathcal{V}_2^t)\\
&{\Lambda}_3^{t+1}={\Lambda}_3^t+\mu(\nabla_{x}\nabla_{z}(f_{\Theta^t}(\mathcal{Z}))-\mathcal{V}_3^t)\\&{\Lambda}_4^{t+1}={\Lambda}_4^t+\mu(\nabla_{y}\nabla_{z}(f_{\Theta^t}(\mathcal{Z}))-\mathcal{V}_4^t).\label{D}
    \end{aligned}
\right.
\end{equation}
\par
The ADMM algorithm is summarized in Algorithm \ref{alg}. 
\begin{table}[t]
\caption{The simulated noisy setting. $\sigma$ denotes the standard deviation. $p$ denotes the sampling rate. $s_1$ and $s_2$ denote the numbers of stripes and deadlines in each corrupted band, respectively. } 
\vspace{-0.3cm}
\begin{center}
\scriptsize
\setlength{\tabcolsep}{1.5pt}
\begin{spacing}{1.05}
\begin{tabular}{ccccccccc}
\toprule
\multirow{3}*{Case}&\multicolumn{2}{c}{Gaussian noise}&\multicolumn{2}{c}{Impulse noise}&\multicolumn{2}{c}{Stripe}&\multicolumn{2}{c}{Deadline}\\
\cmidrule{2-9}
~& \;{added band}& {$\sigma$}& \;{added band}& {$p$}& \;{added band}& {$s_1$}& \;{added band}& {$s_2$}\\
\midrule
{\bf Case 1}&\;{all bands}&$0.2$&\;-\;-&-\;-&\;-\;-&-\;-&\;-\;-&-\;-\\
{\bf Case 2}&\;{all bands}&$0.1$&\;{all bands}&$0.1$&\;-\;-&-\;-&\;-\;-&-\;-\\
{\bf Case 3}&\;{all bands}&$0.1$&\;{all bands}&$0.1$&\;40\%&$[6,15]$&\;-\;-&-\;-\\
{\bf Case 4}&\;{all bands}&$0.1$&\;{all bands}&$0.1$&\;-\;-&-\;-&\;50\%&$[6,10]$\\
{\bf Case 5}&\;{all bands}&$0.1$&\;{all bands}&$0.1$&\;$40\%$&$[6,15]$&\;50\%&$[6,10]$\\
\toprule
\end{tabular}
\vspace{-0.75cm}
\end{spacing}
\end{center}
\label{noisy}
\end{table}
\subsection{The Automatic Stopping Criterion} \label{auto_ter}
We introduce an automatic stopping criterion to demonstrate that the proposed S2DIP can be automatically terminated based on its convergence property. Specifically, we use the relative error
\begin{equation}
{\bf RelErr} = \frac{\lVert f_{\Theta^{t+1}}({\mathcal Z})-f_{\Theta^{t}}({\mathcal Z})\rVert_{F}^2}{\lVert f_{\Theta^{t}}({\mathcal Z})\rVert_{F}^2}\label{RelErr}
\end{equation}
and iteration step to evaluate the convergence degree. Here, $f_{\Theta^{t}}({\mathcal Z})\in{\mathbb R}^{H\times W\times B}$ denotes the $t$-th network output. We set two parameters $r$ and $t_{max}$ as the tolerance of ${\bf RelErr}$ and the maximum number of iteration steps, respectively. The iteration will be terminated if: 1) ${\bf RelErr}$ of the $t$-th iteration is lower than $r$, or 2) the iteration number $t$ exceeds $t_{max}$. Cooperatively, ${\bf RelErr}$ evaluates the convergent degree of the optimization process while $t_{max}$ guarantees the final stopping of the iteration.\par
We use S2DIP* to denote the highest peak signal-to-noise ratio (PSNR) value of the proposed method and use S2DIP to denote the proposed method using the automatic stopping criterion. If the performance of S2DIP is comparable to that of S2DIP*, it means that S2DIP converges to the point with the highest PSNR value, which alleviates the semi-convergence of DIP. It is worth emphasizing that the original DIP could not use such stopping criterion since DIP suffers from semi-convergence. Using such stopping criterion for DIP would result in poor performance, as DIP always converges to a noisy HSI. For convenience, we report the highest PSNR value of DIP in the experiments.
\begin{figure*}[t]
\scriptsize
\setlength{\tabcolsep}{0.9pt}
\begin{center}
\begin{tabular}{ccccccc}
\includegraphics[width=0.14\textwidth]{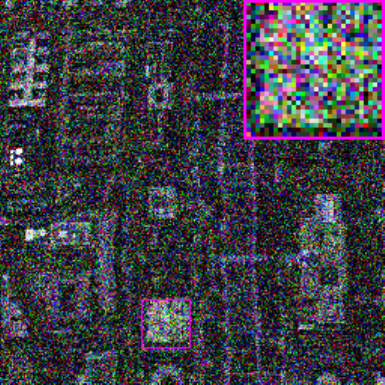}&
\includegraphics[width=0.14\textwidth]{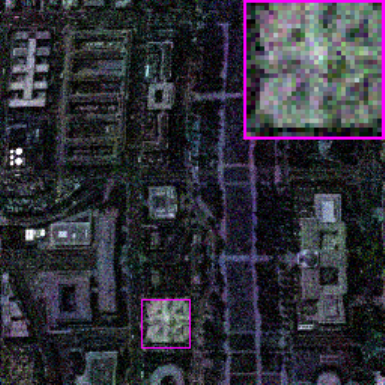}&
\includegraphics[width=0.14\textwidth]{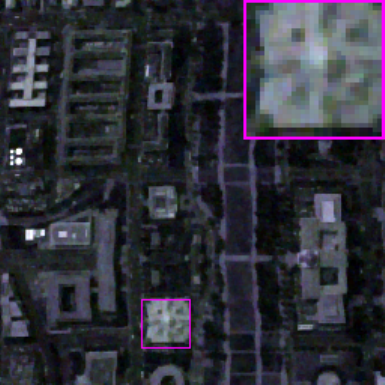}&
\includegraphics[width=0.14\textwidth]{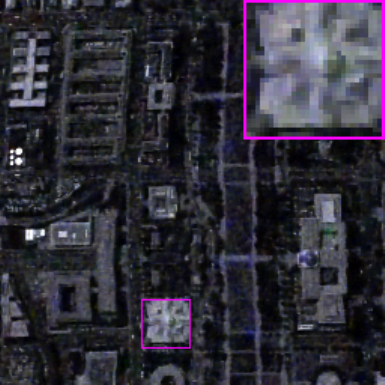}&
\includegraphics[width=0.14\textwidth]{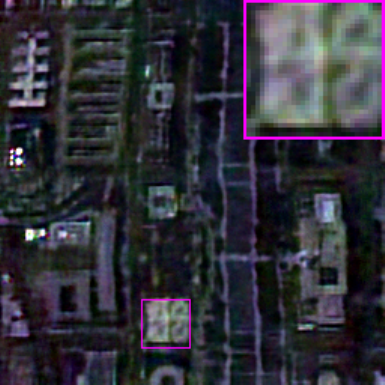}&
\includegraphics[width=0.14\textwidth]{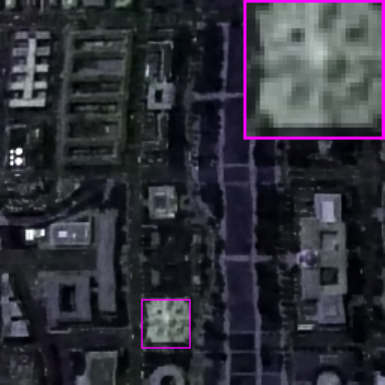}&
\includegraphics[width=0.14\textwidth]{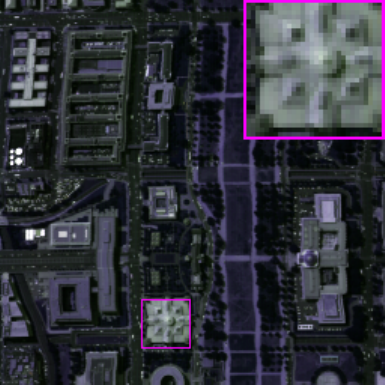}\\
\vspace{0.05cm}
PSNR 16.070 dB&PSNR 26.018 dB&PSNR 30.658 dB&PSNR 31.181 dB&PSNR 27.763 dB&PSNR 31.560 dB&PSNR Inf\\
\includegraphics[width=0.14\textwidth]{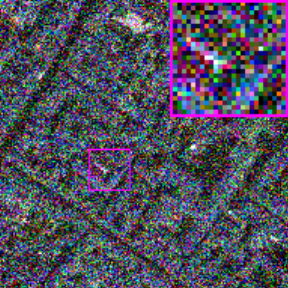}&
\includegraphics[width=0.14\textwidth]{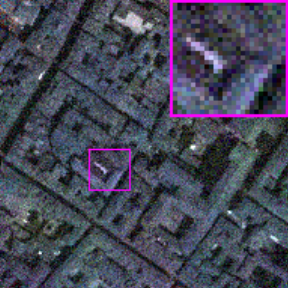}&
\includegraphics[width=0.14\textwidth]{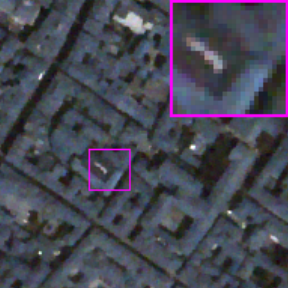}&
\includegraphics[width=0.14\textwidth]{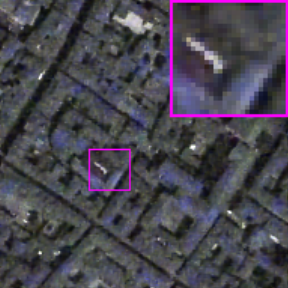}&
\includegraphics[width=0.14\textwidth]{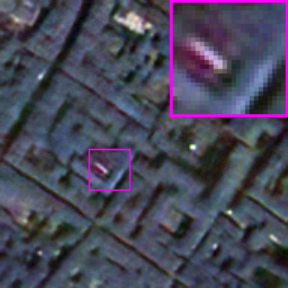}&
\includegraphics[width=0.14\textwidth]{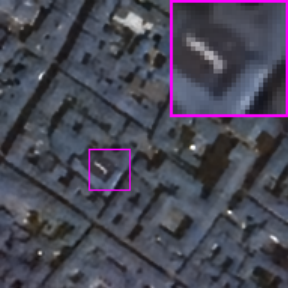}&
\includegraphics[width=0.14\textwidth]{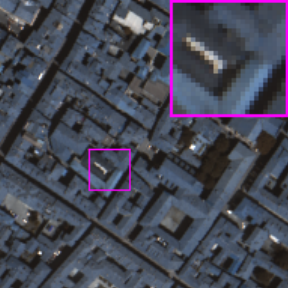}\\
\vspace{0.05cm}
PSNR 15.289 dB&PSNR 25.282 dB&PSNR 30.452 dB&PSNR 29.905 dB&PSNR 27.836 dB&PSNR 31.570 dB&PSNR Inf\\
\includegraphics[width=0.14\textwidth]{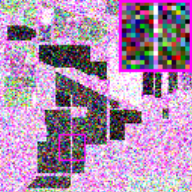}&
\includegraphics[width=0.14\textwidth]{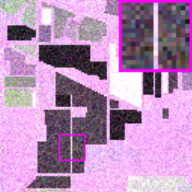}&
\includegraphics[width=0.14\textwidth]{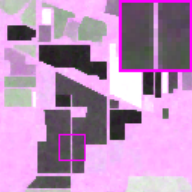}&
\includegraphics[width=0.14\textwidth]{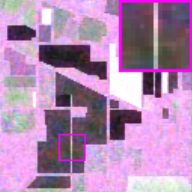}&
\includegraphics[width=0.14\textwidth]{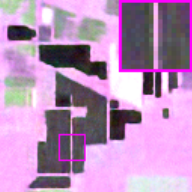}&
\includegraphics[width=0.14\textwidth]{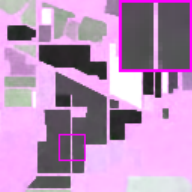}&
\includegraphics[width=0.14\textwidth]{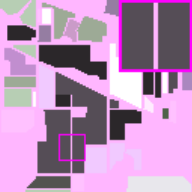}\\
\vspace{0.05cm}
PSNR 14.562 dB&PSNR 24.622 dB&PSNR 31.968 dB&PSNR 26.136 dB&PSNR 27.113 dB&PSNR 32.410 dB&PSNR Inf\\
\includegraphics[width=0.14\textwidth]{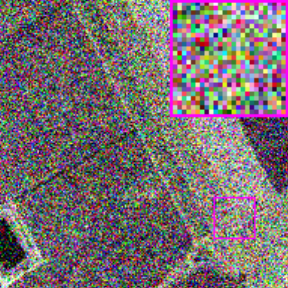}&
\includegraphics[width=0.14\textwidth]{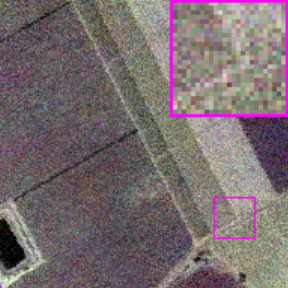}&
\includegraphics[width=0.14\textwidth]{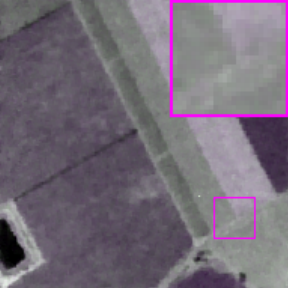}&
\includegraphics[width=0.14\textwidth]{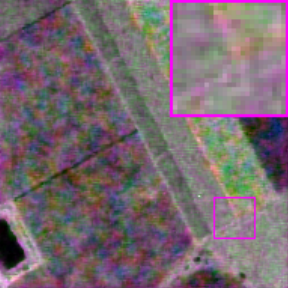}&
\includegraphics[width=0.14\textwidth]{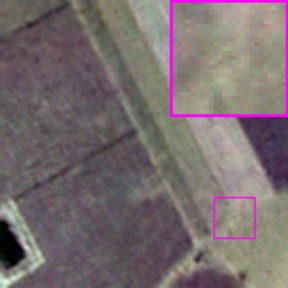}&
\includegraphics[width=0.14\textwidth]{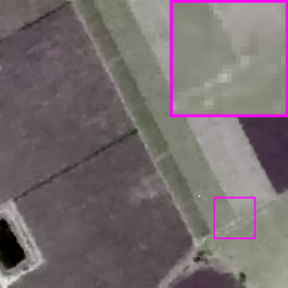}&
\includegraphics[width=0.14\textwidth]{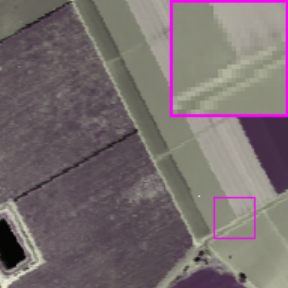}\\
\vspace{0.05cm}
PSNR 15.280 dB&PSNR 26.088 dB&PSNR 33.470 dB&PSNR 24.602 dB&PSNR 31.771 dB&PSNR 35.068 dB&PSNR Inf\\
\vspace{-0.3cm}
Noisy & LRMR\cite{LRMR} & LRTDTV\cite{LRTDTV} &HSID-CNN \cite{HSID-CNN}&DIP 2D\cite{DeepHS} &S2DIP & GT\\
\end{tabular}
\end{center}
\caption{The denoising HSIs by different methods for {\bf Case 1}. Each row from top to down lists {\it WDC mall} consisted of the 5-th, 15-th, and 30-th bands, {\it Pavia} consisted of the 5-th, 15-th, and 30-th bands, {\it Indian pines} consisted of the 15-th, 25-th, and 29-th bands, and {\it Salinas} consisted of the 18-th, 25-th, and 32-th bands.\label{HSI_fig}}
\vspace{-0.3cm}
\end{figure*}
\begin{table*}[!h]
\caption{The quantitative results by different methods on HSIs. The {\bf BEST} value is highlighted by {\bf BOLDFACE}. The \underline{second-best} value is highlighted by \underline{underlined}.\label{HSI_tab}}\vspace{-0.4cm}
\begin{center}
\scriptsize
\setlength{\tabcolsep}{4pt}
\begin{spacing}{1.00}
\begin{tabular}{clccccccccccccccc}
\toprule
\multicolumn{2}{c}{Case}&\multicolumn{3}{c}{\bf Case 1}&\multicolumn{3}{c}{\bf Case 2}&\multicolumn{3}{c}{\bf Case 3}&\multicolumn{3}{c}{\bf Case 4}&\multicolumn{3}{c}{\bf Case 5}\\
\toprule
Data&Method&PSNR &SSIM &SAM\;\;\; &PSNR &SSIM &SAM\;\;\;&PSNR &SSIM &SAM\;\;\;&PSNR &SSIM &SAM\;\;\;&PSNR &SSIM &SAM\\
\midrule
\multirow{7}*{\it WDC mall}
&{LRMR}
& 26.018 & 0.767 & 0.563\;\;\;& 29.869 & 0.887 & 0.335\;\;\;& 27.090 & 0.835 & 0.461\;\;\;& 29.321 & 0.880 & 0.358\;\;\;& 26.658 & 0.828 & 0.487 \\
~&{LRTDTV}
& 30.658 & 0.899 & 0.218\;\;\;& 31.625 & 0.917 & 0.236\;\;\;& 31.433 & 0.905 & 0.271\;\;\;& 30.965 & 0.904 & 0.272\;\;\;& 30.615 & 0.889 & 0.311 \\
~&{HSID-CNN}
& 31.181 & 0.918 & 0.210\;\;\;& 25.222 & 0.792 & 0.234\;\;\;& 23.237 & 0.740 & 0.306\;\;\;& 25.424 & 0.791 & 0.244\;\;\;& 23.391 & 0.737 & 0.313 \\
~&{DIP 2D}
& 27.763 & 0.826 & 0.265\;\;\;& 24.621 & 0.727 & 0.172\;\;\;& 22.753 & 0.642 & 0.240\;\;\;& 24.591 & 0.737 & 0.202\;\;\;& 22.607 & 0.617 & 0.221 \\
~&{DIP 3D}
& 27.784 & 0.828 & 0.264\;\;\;& 24.714 & 0.744 & 0.210\;\;\;& 23.306 & 0.689 & 0.244\;\;\;& 24.553 & 0.731 & 0.199\;\;\;& 23.116 & 0.675 & 0.225 \\
~&{S2DIP}
& \underline{31.560} & \underline{0.920} & \bf{0.112}\;\;\;& \underline{32.503} & \underline{0.931} & \bf{0.102}\;\;\;& \underline{32.333} & \underline{0.930} & \underline{0.112}\;\;\;& \underline{32.585} & \underline{0.932} & \bf{0.102}\;\;\;& \underline{32.004} & \underline{0.924} & \bf{0.111} \\
~&{S2DIP*}
& \bf{31.740} & \bf{0.922} & \underline{0.113}\;\;\;& \bf{32.886} & \bf{0.937} & \underline{0.105}\;\;\;& \bf{32.620} & \bf{0.933} & \bf{0.110}\;\;\;& \bf{32.662} & \bf{0.933} & \underline{0.104}\;\;\;& \bf{32.442} & \bf{0.931} & \underline{0.116} \\
\midrule
\multirow{7}*{\it Pavia}
&{LRMR}
& 25.282 & 0.751 & 0.259\;\;\;& 29.467 & 0.886 & 0.153\;\;\;& 27.040 & 0.836 & 0.240\;\;\;& 28.050 & 0.863 & 0.193\;\;\;& 25.843 & 0.811 & 0.269 \\
~&{LRTDTV}
& 30.452 & 0.901 & 0.103\;\;\;& 31.634 & 0.924 & \underline{0.105}\;\;\;& 31.243 & \underline{0.907} & 0.131\;\;\;& 30.105 & 0.900 & 0.148\;\;\;& 29.417 & 0.878 & 0.175 \\
~&{HSID-CNN}
& 29.905 & 0.906 & 0.114\;\;\;& 26.430 & 0.856 & 0.124\;\;\;& 24.307 & 0.815 & 0.168\;\;\;& 26.513 & 0.842 & 0.146\;\;\;& 24.611 & 0.797 & 0.191 \\
~&{DIP 2D}
& 27.836 & 0.845 & 0.118\;\;\;& 25.472 & 0.808 & 0.112\;\;\;& 23.797 & 0.740 & 0.124\;\;\;& 25.480 & 0.811 & 0.098\;\;\;& 23.738 & 0.740 & 0.128 \\
~&{DIP 3D}
& 27.781 & 0.840 & 0.141\;\;\;& 25.844 & 0.822 & 0.111\;\;\;& 24.239 & 0.782 & 0.109\;\;\;& 25.571 & 0.810 & 0.106\;\;\;& 24.277 & 0.784 & 0.122 \\
~&{S2DIP}
& \underline{31.570} & \underline{0.928} & \underline{0.062}\;\;\;& \underline{32.790} & \underline{0.943} & \bf{0.058}\;\;\;& \underline{32.554} & \bf{0.945} & \underline{0.063}\;\;\;& \underline{32.839} & \underline{0.943} & \underline{0.060}\;\;\;& \underline{32.366} & \underline{0.940} & \underline{0.065} \\
~&{S2DIP*}
& \bf{31.674} & \bf{0.930} & \bf{0.061}\;\;\;& \bf{33.016} & \bf{0.946} & \bf{0.058}\;\;\;& \bf{32.710} & \bf{0.945} & \bf{0.061}\;\;\;& \bf{32.981} & \bf{0.945} & \bf{0.058}\;\;\;& \bf{32.437} & \bf{0.941} & \bf{0.064} \\
\midrule
\multirow{7}*{\it Indian pines}
&{LRMR}
& 24.622 & 0.725 & 0.135\;\;\;& 28.731 & 0.859 & 0.084\;\;\;& 27.712 & 0.844 & 0.108\;\;\;& 23.256 & 0.756 & 0.203\;\;\;& 23.108 & 0.757 & 0.201 \\
~&{LRTDTV}
& 31.968 & 0.952 & 0.050\;\;\;& 33.881 & \underline{0.965} & 0.043\;\;\;& 32.953 & 0.952 & 0.061\;\;\;& 28.071 & 0.894 & 0.129\;\;\;& 27.982 & \underline{0.886} & 0.146 \\
~&{HSID-CNN}
& 26.136 & 0.846 & 0.113\;\;\;& 25.080 & 0.844 & 0.111\;\;\;& 24.217 & 0.833 & 0.124\;\;\;& 21.286 & 0.750 & 0.181\;\;\;& 21.239 & 0.745 & 0.190 \\
~&{DIP 2D}
& 27.113 & 0.882 & 0.088\;\;\;& 26.193 & 0.874 & 0.078\;\;\;& 24.700 & 0.832 & 0.122\;\;\;& 25.886 & 0.867 & 0.077\;\;\;& 24.434 & 0.831 & 0.112 \\
~&{DIP 3D}
& 26.972 & 0.864 & 0.092\;\;\;& 26.333 & 0.863 & 0.076\;\;\;& 25.356 & 0.847 & 0.089\;\;\;& 26.159 & 0.869 & \underline{0.069}\;\;\;& 25.150 & 0.850 & 0.090 \\
~&{S2DIP}
& \underline{32.410} & \underline{0.967} & \underline{0.040}\;\;\;& \underline{35.120} & \bf{0.984} & \underline{0.035}\;\;\;& \underline{34.009} & \underline{0.983} & \underline{0.040}\;\;\;& \underline{34.584} & \underline{0.983} & \bf{0.037}\;\;\;& \underline{33.522} & \bf{0.978} & \underline{0.043} \\
~&{S2DIP*}
& \bf{32.710} & \bf{0.968} & \bf{0.037}\;\;\;& \bf{35.364} & \bf{0.984} & \bf{0.034}\;\;\;& \bf{34.860} & \bf{0.984} & \bf{0.038}\;\;\;& \bf{35.181} & \bf{0.984} & \bf{0.037}\;\;\;& \bf{33.769} & \bf{0.978} & \bf{0.042} \\
\midrule
\multirow{7}*{\it Salinas}
&{LRMR}
& 26.088 & 0.652 & 0.148\;\;\;& 29.786 & 0.808 & 0.098\;\;\;& 27.287 & 0.757 & 0.172\;\;\;& 28.354 & 0.782 & 0.166\;\;\;& 25.422 & 0.717 & 0.228 \\
~&{LRTDTV}
& 33.470 & 0.904 & 0.074\;\;\;& 34.360 & 0.905 & 0.075\;\;\;& 33.471 & 0.887 & 0.091\;\;\;& 32.013 & 0.871 & 0.133\;\;\;& 31.915 & 0.847 & 0.165 \\
~&{HSID-CNN}
& 24.602 & 0.713 & 0.244\;\;\;& 23.273 & 0.673 & 0.266\;\;\;& 22.040 & 0.632 & 0.291\;\;\;& 22.824 & 0.653 & 0.287\;\;\;& 21.270 & 0.596 & 0.317 \\
~&{DIP 2D}
& 31.771 & 0.891 & 0.069\;\;\;& 27.592 & 0.812 & 0.116\;\;\;& 25.885 & 0.787 & 0.152\;\;\;& 27.527 & 0.812 & 0.114\;\;\;& 25.492 & 0.782 & 0.152 \\
~&{DIP 3D}
& 32.729 & 0.916 & 0.058\;\;\;& 28.141 & 0.840 & 0.109\;\;\;& 25.983 & 0.807 & 0.134\;\;\;& 28.211 & 0.846 & 0.104\;\;\;& 26.185 & 0.806 & \underline{0.141} \\
~&{S2DIP}
& \underline{35.068} & \underline{0.945} & \bf{0.039}\;\;\;& \underline{35.997} & \underline{0.954} & \bf{0.036}\;\;\;& \underline{35.123} & \underline{0.950} & \underline{0.042}\;\;\;& \underline{36.056} & \underline{0.956} & \bf{0.034}\;\;\;& \underline{35.084} & \underline{0.948} & \bf{0.043} \\
~&{S2DIP*}
& \bf{35.224} & \bf{0.946} & \underline{0.041}\;\;\;& \bf{36.223} & \bf{0.956} & \underline{0.037}\;\;\;& \bf{35.474} & \bf{0.952} & \bf{0.041}\;\;\;& \bf{36.212} & \bf{0.957} & \underline{0.036}\;\;\;& \bf{35.259} & \bf{0.951} & \bf{0.043} \\
\bottomrule
\end{tabular}
\end{spacing}
\end{center}\vspace{-0.6cm}
\end{table*}
\begin{figure*}[t]
\scriptsize
\setlength{\tabcolsep}{0.9pt}
\begin{center}
\begin{tabular}{ccccccc}
\includegraphics[width=0.14\textwidth]{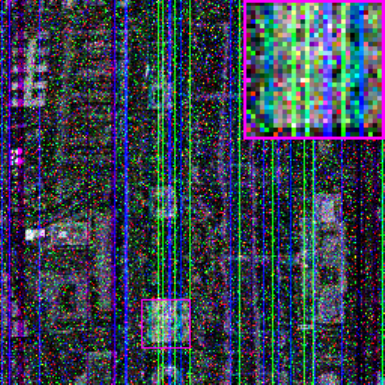}&
\includegraphics[width=0.14\textwidth]{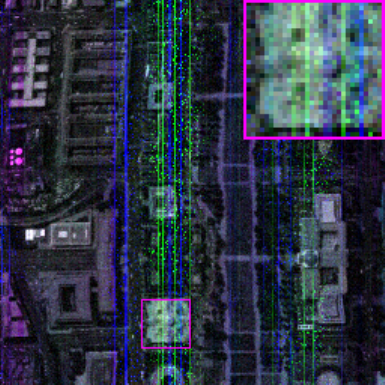}&
\includegraphics[width=0.14\textwidth]{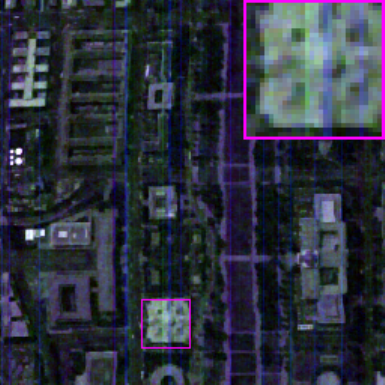}&
\includegraphics[width=0.14\textwidth]{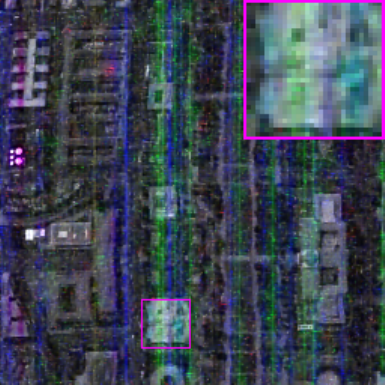}&
\includegraphics[width=0.14\textwidth]{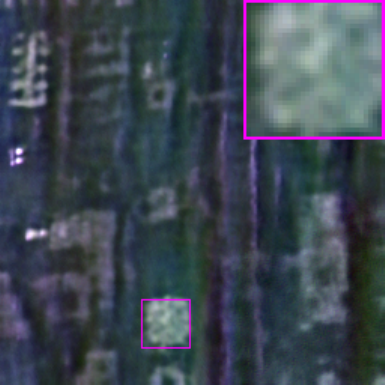}&
\includegraphics[width=0.14\textwidth]{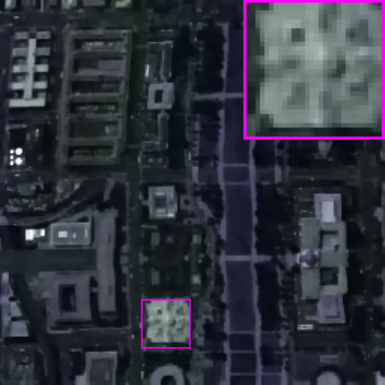}&
\includegraphics[width=0.14\textwidth]{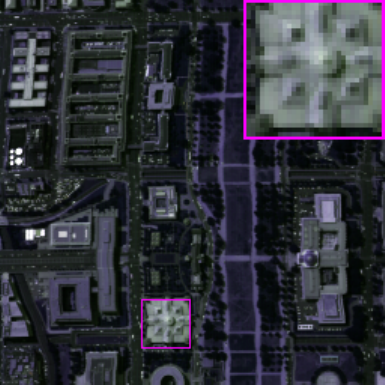}\\
\vspace{0.05cm}
PSNR 13.198 dB&PSNR 26.658 dB&PSNR 30.615 dB&PSNR 23.391 dB&PSNR 22.607 dB&PSNR 32.004 dB&PSNR Inf\\
\includegraphics[width=0.14\textwidth]{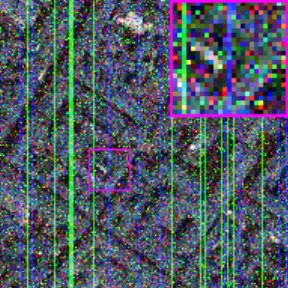}&
\includegraphics[width=0.14\textwidth]{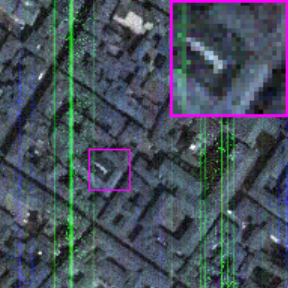}&
\includegraphics[width=0.14\textwidth]{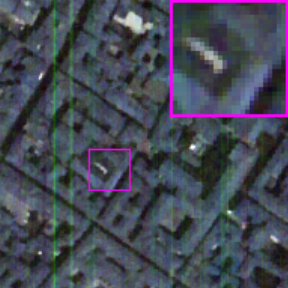}&
\includegraphics[width=0.14\textwidth]{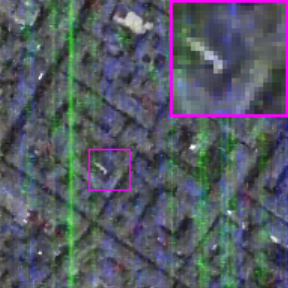}&
\includegraphics[width=0.14\textwidth]{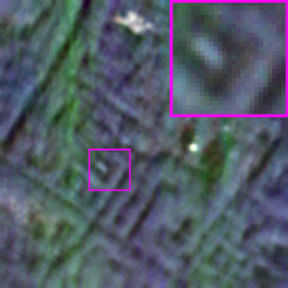}&
\includegraphics[width=0.14\textwidth]{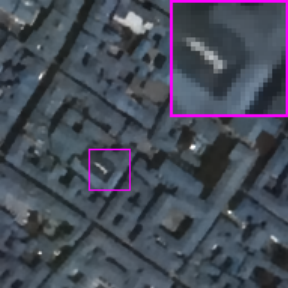}&
\includegraphics[width=0.14\textwidth]{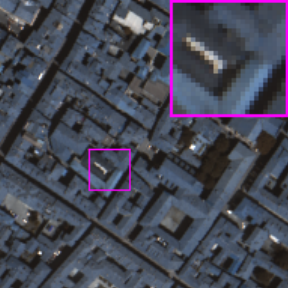}\\
\vspace{0.05cm}
PSNR 13.603 dB&PSNR 25.843 dB&PSNR 29.417 dB&PSNR 24.611 dB&PSNR 23.738 dB&PSNR 32.366 dB&PSNR Inf\\
\includegraphics[width=0.14\textwidth]{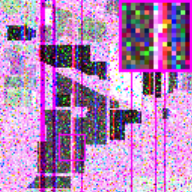}&
\includegraphics[width=0.14\textwidth]{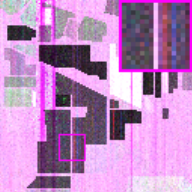}&
\includegraphics[width=0.14\textwidth]{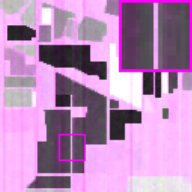}&
\includegraphics[width=0.14\textwidth]{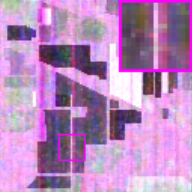}&
\includegraphics[width=0.14\textwidth]{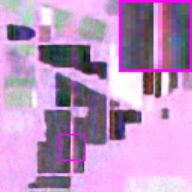}&
\includegraphics[width=0.14\textwidth]{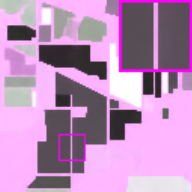}&
\includegraphics[width=0.14\textwidth]{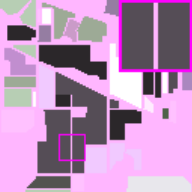}\\
\vspace{0.05cm}
PSNR 13.480 dB&PSNR 23.108 dB&PSNR 27.982 dB&PSNR 21.239 dB&PSNR 24.434 dB&PSNR 33.522 dB&PSNR Inf\\
\includegraphics[width=0.14\textwidth]{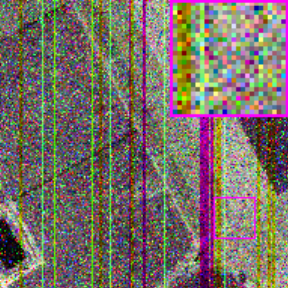}&
\includegraphics[width=0.14\textwidth]{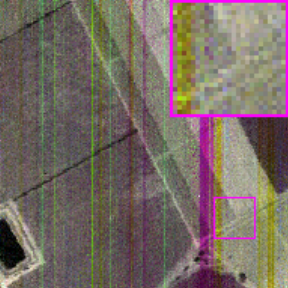}&
\includegraphics[width=0.14\textwidth]{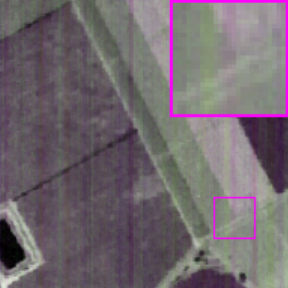}&
\includegraphics[width=0.14\textwidth]{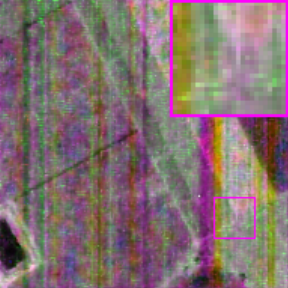}&
\includegraphics[width=0.14\textwidth]{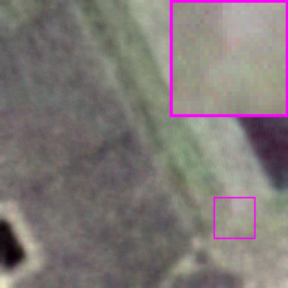}&
\includegraphics[width=0.14\textwidth]{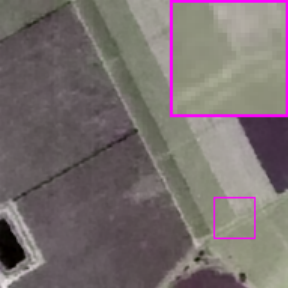}&
\includegraphics[width=0.14\textwidth]{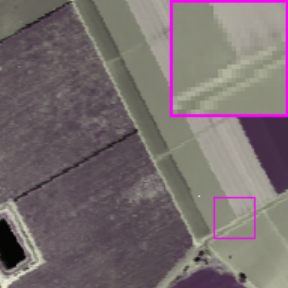}\\
\vspace{0.05cm}
PSNR 13.372 dB&PSNR 25.422 dB&PSNR 31.915 dB&PSNR 21.270 dB&PSNR 25.492 dB&PSNR 35.084 dB&PSNR Inf\\
\vspace{-0.3cm}
Noisy & LRMR\cite{LRMR} & LRTDTV\cite{LRTDTV} &HSID-CNN \cite{HSID-CNN}&DIP 2D\cite{DeepHS} &S2DIP &GT\\
\end{tabular}
\end{center}
\caption{The denoising HSIs by different methods for {\bf Case 5}. Each row from top to down lists {\it WDC mall} consisted of the 5-th, 15-th, and 30-th bands, {\it Pavia} consisted of the 5-th, 15-th, and 30-th bands, {\it Indian pines} consisted of the 15-th, 25-th, and 29-th bands, and {\it Salinas} consisted of the 18-th, 25-th, and 32-th bands.\label{HSI_fig_2}}
\vspace{-0.2cm}
\end{figure*}
\begin{figure*}[t]
\scriptsize
\setlength{\tabcolsep}{0.9pt}
\begin{center}
\begin{tabular}{c}
\vspace{-0.3cm}
\includegraphics[width=1\textwidth]{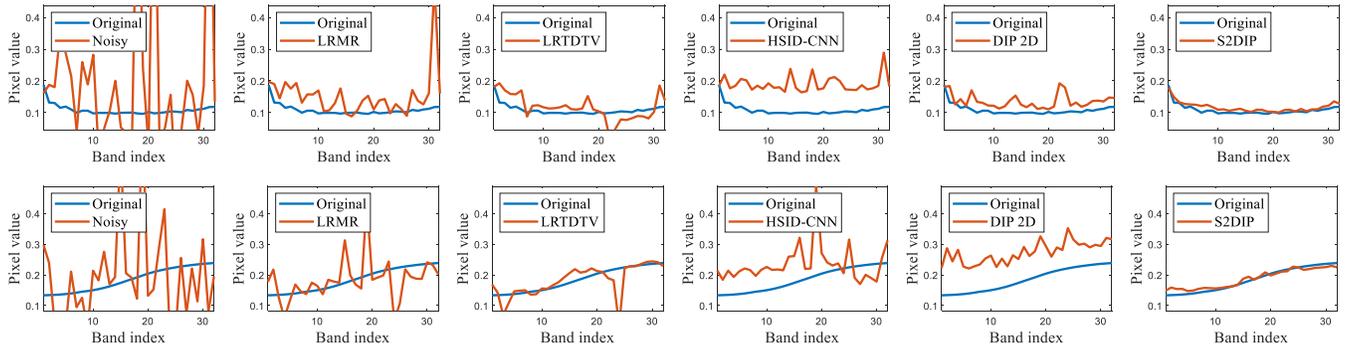}\\
\end{tabular}
\end{center}
\caption{The spectral curves of the recovered results by different methods on HSIs {\it WDC mall} and {\it Pavia} for {\bf Case 5}.\label{HSI_spec}}
\vspace{-0.2cm}
\end{figure*}
\begin{figure*}[t]
\scriptsize
\setlength{\tabcolsep}{0.9pt}
\begin{center}
\begin{tabular}{c}
\vspace{-0.3cm}
\includegraphics[width=1\textwidth]{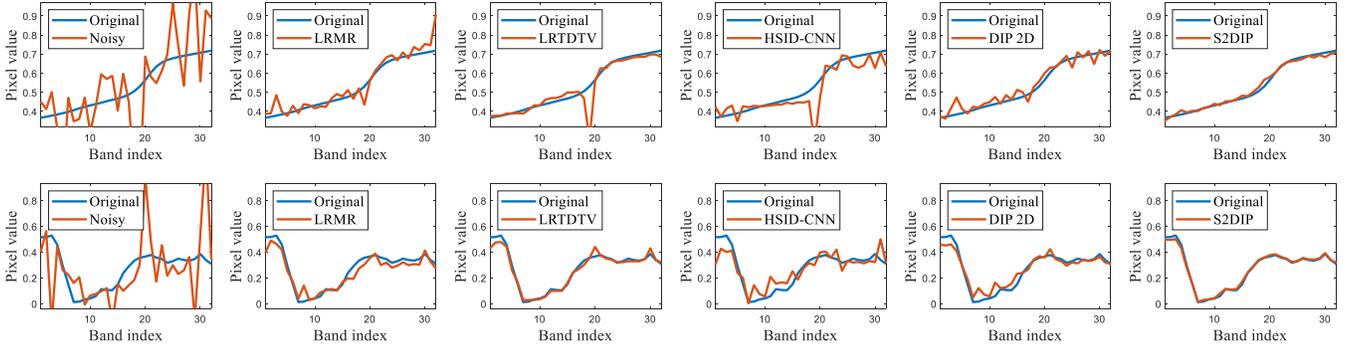}\\
\end{tabular}
\end{center}
\caption{The spectral curves of the recovered results by different methods on HSIs {\it Indian pines} and {\it Salinas} for {\bf Case 5}.\label{HSI_spec_2}}
\vspace{-0.2cm}
\end{figure*}
\begin{table*}[!h]
\caption{The quantitative results by different methods on MSIs. The {\bf BEST} value is highlighted by {\bf BOLDFACE}. The \underline{second-best} value is highlighted by \underline{underlined}.\label{MSI_tab}}\vspace{-0.4cm}
\begin{center}
\scriptsize
\setlength{\tabcolsep}{4pt}
\begin{spacing}{1.00}
\begin{tabular}{clccccccccccccccc}
\toprule
\multicolumn{2}{c}{Case}&\multicolumn{3}{c}{\bf Case 1}&\multicolumn{3}{c}{\bf Case 2}&\multicolumn{3}{c}{\bf Case 3}&\multicolumn{3}{c}{\bf Case 4}&\multicolumn{3}{c}{\bf Case 5}\\
\toprule
Data&Method&PSNR &SSIM &SAM\;\;\; &PSNR &SSIM &SAM\;\;\;&PSNR &SSIM &SAM\;\;\;&PSNR &SSIM &SAM\;\;\;&PSNR &SSIM &SAM\\
\midrule
\multirow{7}*{\it Trash}
&{LRMR}
& 25.437 & 0.642 & 0.302\;\;\;& 29.548 & 0.820 & 0.170\;\;\;& 27.210 & 0.780 & 0.261\;\;\;& 27.898 & 0.804 & 0.202\;\;\;& 25.714 & 0.753 & 0.304 \\
~&{LRTDTV}
& 29.812 & 0.922 & 0.145\;\;\;& 32.088 & {0.932} & 0.130\;\;\;& 31.300 & 0.916 & 0.154\;\;\;& 30.939 & \underline{0.915} & 0.157\;\;\;& 30.377 & 0.903 & 0.176 \\
~&{HSID-CNN}
& 28.389 & 0.857 & 0.168\;\;\;& 25.111 & 0.785 & 0.183\;\;\;& 23.727 & 0.749 & 0.219\;\;\;& 24.857 & 0.776 & 0.196\;\;\;& 23.325 & 0.726 & 0.235 \\
~&{DIP 2D}
& 31.291 & 0.922 & \underline{0.105}\;\;\;& 26.985 & 0.865 & 0.131\;\;\;& 25.530 & 0.850 & 0.145\;\;\;& 27.145 & 0.877 & \underline{0.109}\;\;\;& 25.211 & 0.846 & 0.142 \\
~&{DIP 3D}
& 29.895 & 0.931 & 0.134\;\;\;& 26.909 & 0.901 & 0.129\;\;\;& 24.979 & 0.873 & 0.141\;\;\;& 26.806 & 0.891 & 0.137\;\;\;& 24.820 & 0.861 & 0.155 \\
~&{S2DIP}
& \underline{33.352} & \underline{0.971} & \bf{0.078}\;\;\;& \underline{35.549} & \underline{0.978} & \underline{0.064}\;\;\;& \underline{34.891} & \underline{0.976} & \bf{0.068}\;\;\;& \underline{35.488} & \bf{0.978} & \bf{0.064}\;\;\;& \underline{34.694} & \underline{0.975} & \underline{0.073} \\
~&{S2DIP*}
& \bf{33.537} & \bf{0.972} & \bf{0.078}\;\;\;& \bf{35.716} & \bf{0.978} & \bf{0.062}\;\;\;& \bf{35.112} & \bf{0.977} & \underline{0.069}\;\;\;& \bf{35.690} & \bf{0.978} & \bf{0.064}\;\;\;& \bf{34.985} & \bf{0.976} & \bf{0.071} \\
\midrule
\multirow{7}*{\it Borad}
&{LRMR}
& 25.610 & 0.687 & 0.333\;\;\;& 29.552 & 0.841 & 0.191\;\;\;& 27.256 & 0.803 & 0.274\;\;\;& 27.596 & 0.820 & 0.227\;\;\;& 25.622 & 0.778 & 0.330 \\
~&{LRTDTV}
& 28.938 & \underline{0.904} & 0.137\;\;\;& 31.032 & 0.917 & 0.150\;\;\;& 30.645 & \underline{0.908} & 0.166\;\;\;& 30.150 & 0.901 & 0.179\;\;\;& 29.630 & 0.884 & 0.218 \\
~&{HSID-CNN}
& 28.019 & 0.856 & 0.178\;\;\;& 24.984 & 0.785 & 0.188\;\;\;& 23.312 & 0.743 & 0.220\;\;\;& 24.444 & 0.769 & 0.208\;\;\;& 23.031 & 0.729 & 0.247 \\
~&{DIP 2D}
& 30.258 & 0.888 & 0.136\;\;\;& 26.765 & 0.842 & 0.134\;\;\;& 25.226 & 0.814 & 0.138\;\;\;& 26.886 & 0.840 & \underline{0.137}\;\;\;& 25.167 & 0.811 & 0.160 \\
~&{DIP 3D}
& 29.768 & 0.899 & 0.145\;\;\;& 26.886 & 0.859 & 0.149\;\;\;& 25.297 & 0.834 & 0.156\;\;\;& 26.562 & 0.860 & \underline{0.137}\;\;\;& 25.225 & 0.833 & 0.164 \\
~&{S2DIP}
& \underline{31.811} & \bf{0.939} & \bf{0.102}\;\;\;& \underline{33.600} & \underline{0.956} & \bf{0.085}\;\;\;& \underline{33.272} & \bf{0.955} & \underline{0.091}\;\;\;& \underline{33.390} & \underline{0.956} & \bf{0.085}\;\;\;& \underline{33.188} & \underline{0.953} & \bf{0.090} \\
~&{S2DIP*}
& \bf{31.925} & \bf{0.939} & \underline{0.103}\;\;\;& \bf{33.712} & \bf{0.957} & \underline{0.086}\;\;\;& \bf{33.383} & \bf{0.955} & \bf{0.090}\;\;\;& \bf{33.634} & \bf{0.957} & \bf{0.085}\;\;\;& \bf{33.289} & \bf{0.955} & \underline{0.091} \\
\midrule
\multirow{7}*{\it Balloons}
&{LRMR}
& 28.311 & 0.794 & 0.223\;\;\;& 27.891 & 0.734 & 0.291\;\;\;& 27.618 & 0.758 & 0.299\;\;\;& 26.801 & 0.729 & 0.315\;\;\;& 28.938 & 0.801 & 0.254 \\
~&{LRTDTV}
& 33.928 & 0.949 & 0.235\;\;\;& 35.406 & 0.948 & 0.193\;\;\;& 34.788 & \underline{0.937} & 0.203\;\;\;& 34.359 & 0.944 & 0.210\;\;\;& 34.154 & 0.933 & 0.226 \\
~&{HSID-CNN}
& 28.452 & 0.839 & 0.317\;\;\;& 24.586 & 0.680 & 0.338\;\;\;& 22.761 & 0.625 & 0.366\;\;\;& 24.130 & 0.669 & 0.353\;\;\;& 22.659 & 0.624 & 0.408 \\
~&{DIP 2D}
& 33.989 & 0.948 & 0.189\;\;\;& 27.114 & 0.822 & 0.224\;\;\;& 25.204 & 0.786 & 0.262\;\;\;& 27.182 & 0.818 & 0.239\;\;\;& 25.450 & 0.791 & 0.283 \\
~&{DIP 3D}
& 33.343 & 0.940 & 0.217\;\;\;& 27.257 & 0.817 & 0.242\;\;\;& 25.110 & 0.773 & 0.254\;\;\;& 27.033 & 0.824 & 0.241\;\;\;& 25.319 & 0.786 & 0.275 \\
~&{S2DIP}
& \underline{36.610} & \underline{0.978} & \underline{0.117}\;\;\;& \underline{38.482} & \underline{0.980} & \underline{0.101}\;\;\;& \underline{37.888} & \bf{0.976} & \bf{0.113}\;\;\;& \underline{38.216} & \underline{0.978} & \underline{0.104}\;\;\;& \underline{37.367} & \underline{0.971} & \underline{0.125} \\
~&{S2DIP*}
& \bf{37.435} & \bf{0.980} & \bf{0.115}\;\;\;& \bf{39.152} & \bf{0.985} & \bf{0.094}\;\;\;& \bf{38.082} & \bf{0.976} & \underline{0.114}\;\;\;& \bf{38.828} & \bf{0.983} & \bf{0.100}\;\;\;& \bf{37.881} & \bf{0.973} & \bf{0.121} \\
\midrule
\multirow{7}*{\it Cups}
&{LRMR}
& 27.042 & 0.752 & 0.071\;\;\;& 28.417 & 0.770 & 0.111\;\;\;& 29.147 & 0.812 & 0.093\;\;\;& 26.023 & 0.734 & 0.159\;\;\;& 28.783 & 0.834 & 0.092 \\
~&{LRTDTV}
& 33.804 & \underline{0.963} & 0.079\;\;\;& 35.908 & \underline{0.961} & 0.064\;\;\;& 35.601 & \underline{0.959} & 0.066\;\;\;& 34.547 & \underline{0.950} & 0.076\;\;\;& 33.867 & \underline{0.942} & 0.082 \\
~&{HSID-CNN}
& 28.739 & 0.854 & 0.109\;\;\;& 26.626 & 0.827 & 0.124\;\;\;& 25.617 & 0.809 & 0.140\;\;\;& 25.418 & 0.795 & 0.155\;\;\;& 24.483 & 0.770 & 0.175 \\
~&{DIP 2D}
& 32.473 & 0.942 & 0.076\;\;\;& 29.810 & 0.930 & 0.075\;\;\;& 27.709 & 0.916 & 0.098\;\;\;& 28.963 & 0.926 & {0.072}\;\;\;& 27.801 & 0.912 & 0.106 \\
~&{DIP 3D}
& 32.254 & 0.940 & 0.074\;\;\;& 29.403 & 0.929 & 0.075\;\;\;& 28.183 & 0.910 & 0.093\;\;\;& 29.440 & 0.927 & 0.077\;\;\;& 27.844 & 0.920 & \underline{0.081} \\
~&{S2DIP}
& \underline{36.376} & \bf{0.983} & \bf{0.036}\;\;\;& \underline{38.560} & \bf{0.989} & \bf{0.030}\;\;\;& \underline{37.882} & \bf{0.988} & \underline{0.035}\;\;\;& \underline{38.169} & \bf{0.989} & \bf{0.031}\;\;\;& \underline{37.834} & \bf{0.988} & \bf{0.034} \\
~&{S2DIP*}
& \bf{36.469} & \bf{0.983} & \underline{0.037}\;\;\;& \bf{38.766} & \bf{0.989} & \underline{0.031}\;\;\;& \bf{38.160} & \bf{0.988} & \bf{0.033}\;\;\;& \bf{38.605} & \bf{0.989} & \underline{0.032}\;\;\;& \bf{38.147} & \bf{0.988} & \bf{0.034} \\
\bottomrule
\end{tabular}
\end{spacing}
\end{center}\vspace{-0.5cm}
\end{table*}
\begin{figure*}[t]
\scriptsize
\setlength{\tabcolsep}{0.9pt}
\begin{center}
\begin{tabular}{ccccccc}
\includegraphics[width=0.14\textwidth]{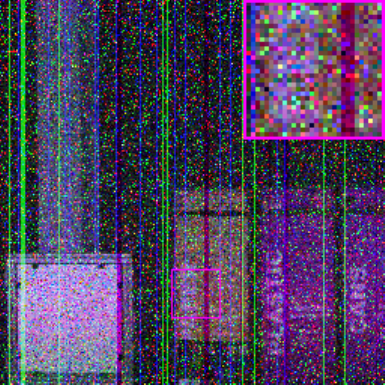}&
\includegraphics[width=0.14\textwidth]{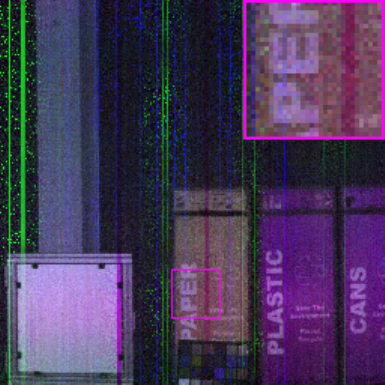}&
\includegraphics[width=0.14\textwidth]{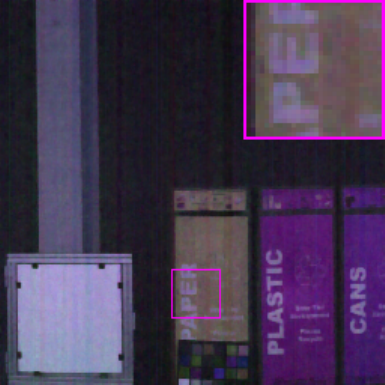}&
\includegraphics[width=0.14\textwidth]{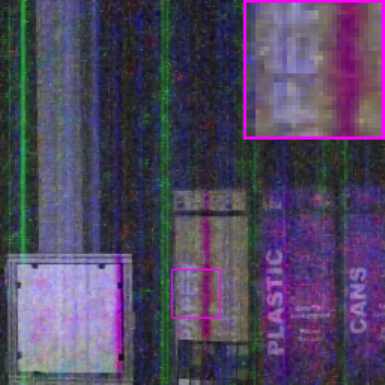}&
\includegraphics[width=0.14\textwidth]{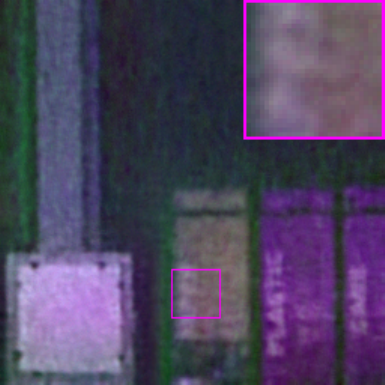}&
\includegraphics[width=0.14\textwidth]{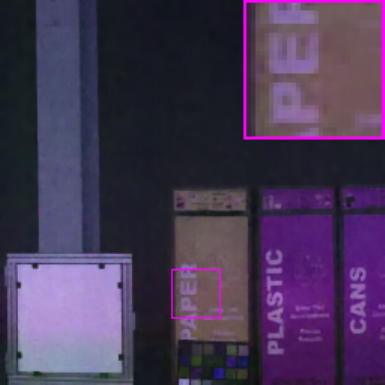}&
\includegraphics[width=0.14\textwidth]{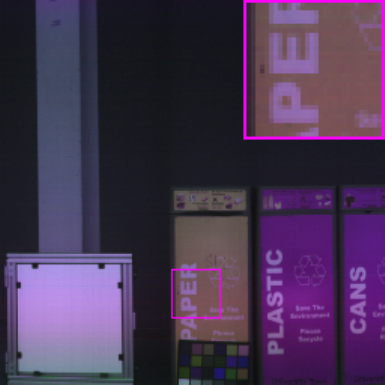}\\
\vspace{0.05cm}
PSNR 13.469 dB&PSNR 25.714 dB&PSNR 30.377 dB&PSNR 23.325 dB&PSNR 25.211 dB&PSNR 34.694 dB&PSNR Inf\\
\includegraphics[width=0.14\textwidth]{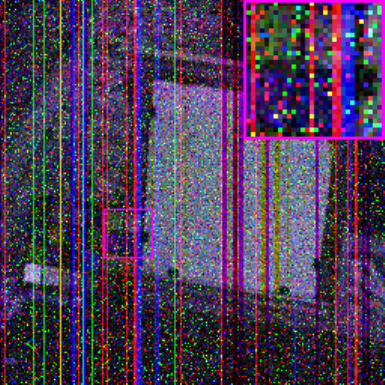}&
\includegraphics[width=0.14\textwidth]{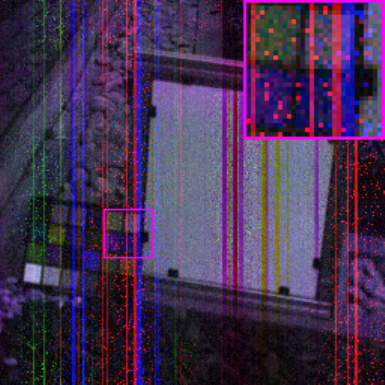}&
\includegraphics[width=0.14\textwidth]{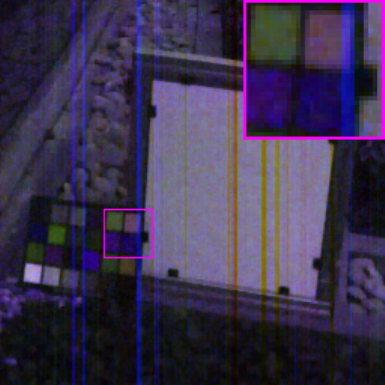}&
\includegraphics[width=0.14\textwidth]{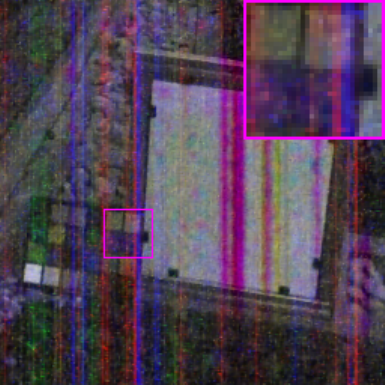}&
\includegraphics[width=0.14\textwidth]{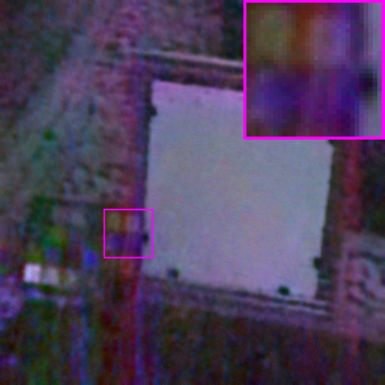}&
\includegraphics[width=0.14\textwidth]{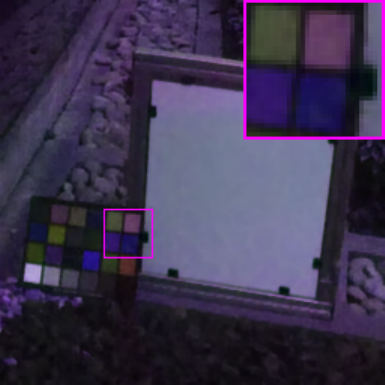}&
\includegraphics[width=0.14\textwidth]{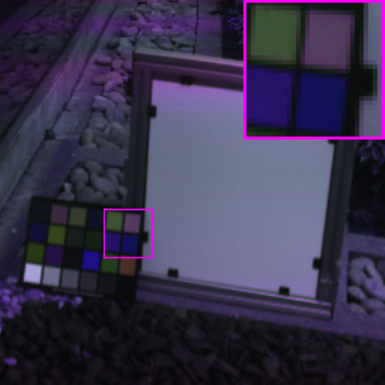}\\
\vspace{0.05cm}
PSNR 13.501 dB&PSNR 25.622 dB&PSNR 29.630 dB&PSNR 23.031 dB&PSNR 25.167 dB&PSNR 33.188 dB&PSNR Inf\\
\includegraphics[width=0.14\textwidth]{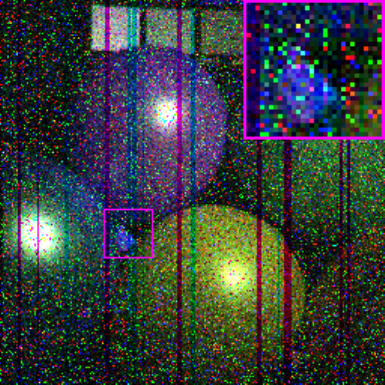}&
\includegraphics[width=0.14\textwidth]{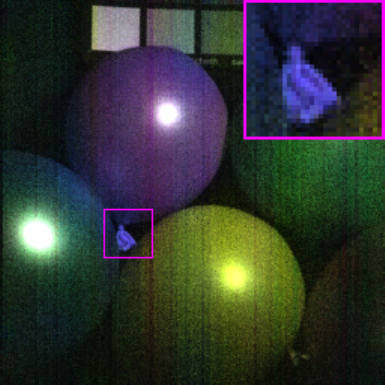}&
\includegraphics[width=0.14\textwidth]{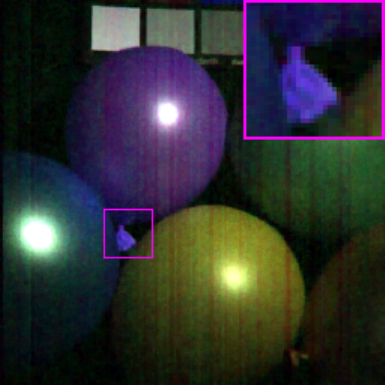}&
\includegraphics[width=0.14\textwidth]{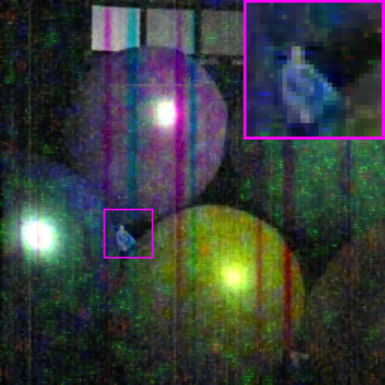}&
\includegraphics[width=0.14\textwidth]{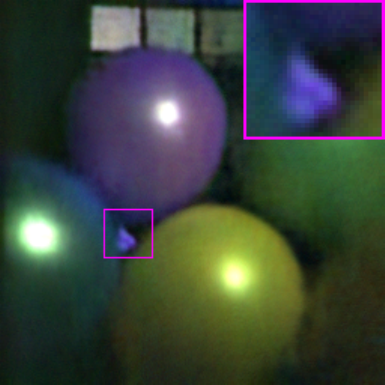}&
\includegraphics[width=0.14\textwidth]{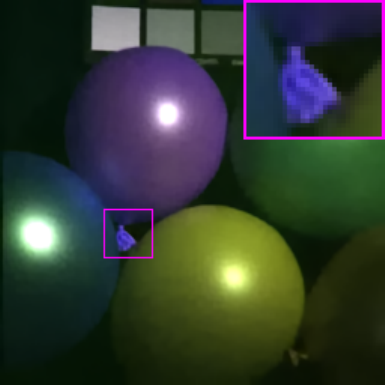}&
\includegraphics[width=0.14\textwidth]{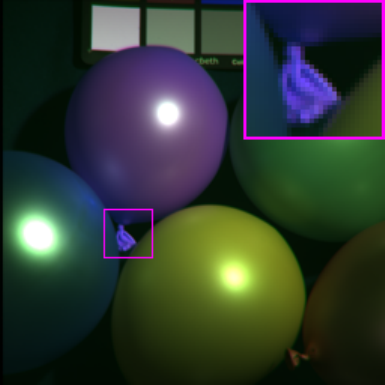}\\
\vspace{0.05cm}
PSNR 13.204 dB&PSNR 28.938 dB&PSNR 34.154 dB&PSNR 22.659 dB&PSNR 25.450 dB&PSNR 37.367 dB&PSNR Inf\\
\includegraphics[width=0.14\textwidth]{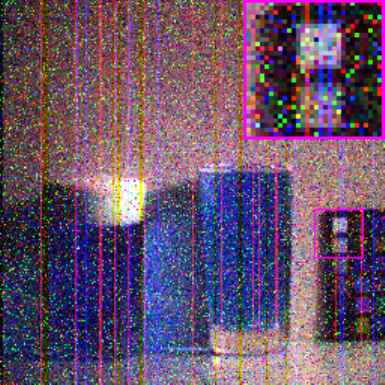}&
\includegraphics[width=0.14\textwidth]{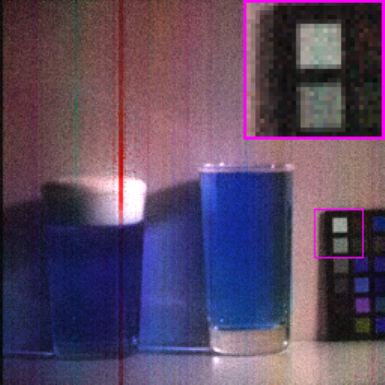}&
\includegraphics[width=0.14\textwidth]{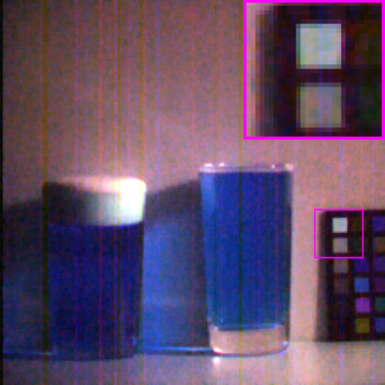}&
\includegraphics[width=0.14\textwidth]{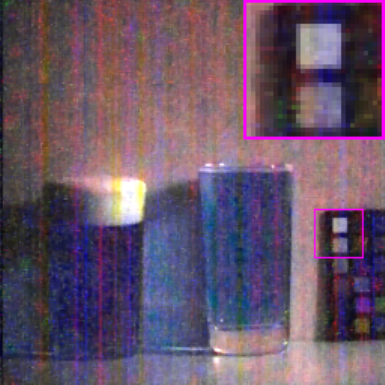}&
\includegraphics[width=0.14\textwidth]{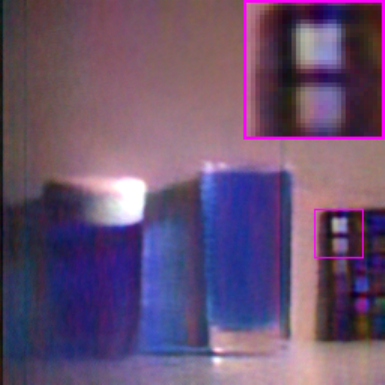}&
\includegraphics[width=0.14\textwidth]{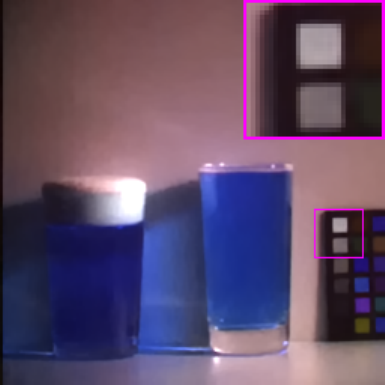}&
\includegraphics[width=0.14\textwidth]{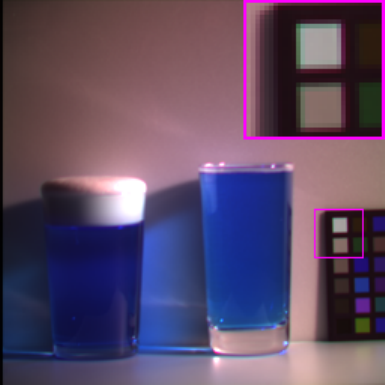}\\
\vspace{0.05cm}
PSNR 13.934 dB&PSNR 28.783 dB&PSNR 33.867 dB&PSNR 24.483 dB&PSNR 27.801 dB&PSNR 37.834 dB&PSNR Inf\\
\vspace{-0.3cm}
Noisy & LRMR\cite{LRMR} & LRTDTV\cite{LRTDTV} &HSID-CNN \cite{HSID-CNN}&DIP 2D\cite{DeepHS} & S2DIP &GT\\
\end{tabular}
\end{center}
\caption{The denoising MSIs by different methods for {\bf Case 5}. Each row from top to down lists {\it Trash} consisted of the 5-th, 15-th, and 25-th bands, {\it Borad} consisted of the 5-th, 15-th, and 25-th bands, {\it Balloons} consisted of the 1-st, 10-th, and 30-th bands, and {\it Cups} consisted of the 1-st, 10-th, and 30-th bands.\label{MSI_fig}}
\end{figure*}
\begin{figure*}[t]
\scriptsize
\setlength{\tabcolsep}{0.9pt}
\begin{center}
\begin{tabular}{c}
\vspace{-0.3cm}
\includegraphics[width=1\textwidth]{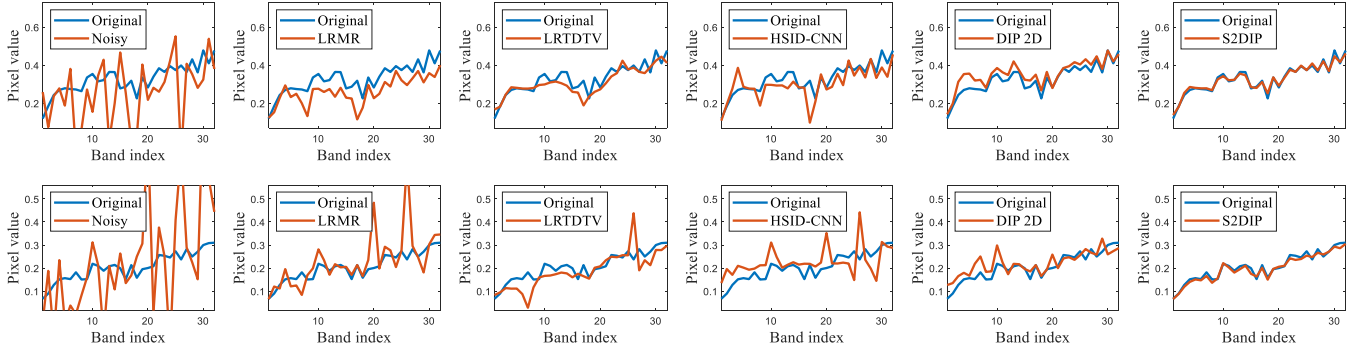}\\
\end{tabular}
\end{center}
\caption{The spectral curves of the recovered results by different methods on MSIs {\it Trash} and {\it Borad} for {\bf Case 5}.\label{MSI_spec}}
\vspace{-0.3cm}
\end{figure*}
\begin{figure*}[t]
\scriptsize
\setlength{\tabcolsep}{0.9pt}
\begin{center}
\begin{tabular}{c}
\vspace{-0.3cm}
\includegraphics[width=1\textwidth]{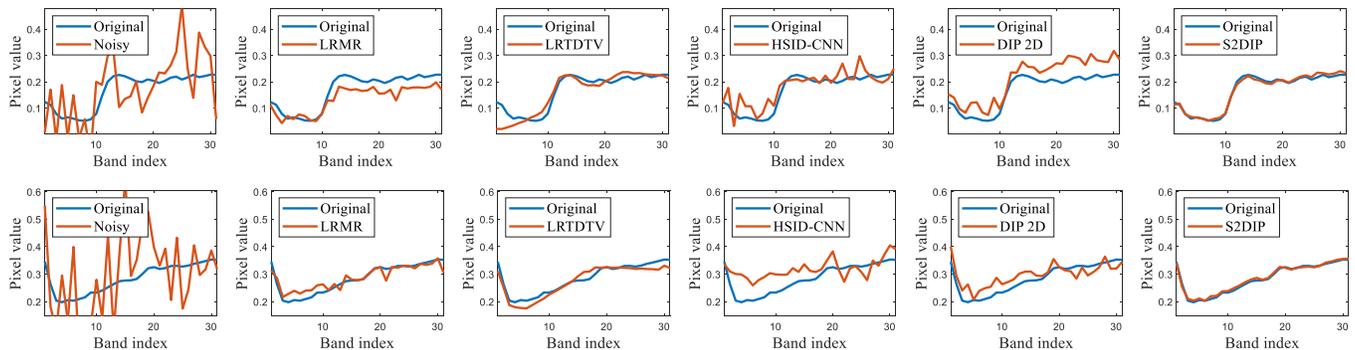}\\
\end{tabular}
\end{center}
\caption{The spectral curves of the recovered results by different methods on MSIs {\it Balloons} and {\it Cups} for {\bf Case 5}.\label{MSI_spec_2}}
\end{figure*}
\section{Experiments} \label{sec:Exp}
Both simulation experiments and real data experiments are involved to verify the validity of the proposed method. The competing methods include the matrix low-rankness-based method LRMR \cite{LRMR}, the tensor low-rankness and total variation-based method LRTDTV \cite{LRTDTV}, the supervised deep CNN-based method HSID-CNN\cite{HSID-CNN}, and the unsupervised DIP methods that based on 2D convolution and 3D convolution (termed as DIP 2D and DIP 3D) \cite{DIP,DeepHS}. The hyperparameters of model-based methods are tuned based on their best PSNR values. The results of HSID-CNN are obtained by using the pre-trained model provided by the author.\par
For simulation experiments, we adopt sub-images of HSIs {\it WDC mall}, {\it Pavia}, {\it Indian pines}, and {\it Salinas}\footnote{\url{http://www.ehu.eus/ccwintco/index.php?title=Hyperspectral Remote Sensing Scenes}} with size $256\times256\times32$, $192\times192\times32$, $128\times128\times32$, and $192\times192\times32$, respectively, to generate simulated noisy HSIs. Additionally, multispectral images (MSIs) including {\it Trash}, {\it Board}\footnote{\url{https://sites.google.com/site/hyperspectralcolorimaging/dataset/general-scenes}}, {\it Balloons}, and {\it Cups}\footnote{\url{https://www.cs.columbia. edu/CAVE/databases/multispectral/}} \cite{cave} are adopted to verify the generalization ability of our method. The size of {\it Trash} and {\it Board} is $256\times256\times32$ and the size of the {\it Ballons} and {\it Cups} is $256\times256\times31$. For all selected images, five noisy cases are established, see Table \ref{noisy}. Based on the noisy setting in Table \ref{noisy}, all noise is randomly performed on the ground-truth (GT) images. For cases with dealines, we change the fidelity term of our method to $\lVert {\mathcal M}\odot({\mathcal Y}-{\mathcal X})-{\mathcal S}\rVert_{F}^2$, where ${\mathcal M}$ denotes the mask whose entries on the positions of deadlines are set to $0$ and other entries are set to $1$, i.e., the removal of deadlines can be seen as an inpainting problem. The inpainting problem can be well addressed by the deep prior of the CNN \cite{DIP,DeepHS}. Here, $\odot$ denotes the element-wise product. We empirically set the tolerance of ${\bf RelErr}$ (i.e., $r$) as 0.01, and ${t}_{max}$ as 7000, which provides good termination points. In addition, we set our hyperparameters $\alpha_1 = 0.1,\alpha_2 = 0.1$, and $\alpha_3=0.01$ for {\bf Case 2-5}. For {\bf Case 1}, we change $\alpha_3=10$ and other parameters remain the same. 
\par
In simulation experiments, the denoising results are numerically evaluated by PSNR, structure similarity (SSIM), and spectral angle mapper (SAM) \cite{sam}. It is worth noting that higher PSNR and SSIM values represent better performance while lower SAM values represent better performance.\par
In real experiments, noisy HSIs {\it Urban} of size $288\times288\times210$ with its band 134 to band 165 and {\it Indian}\footnote{\url{https://purr.purdue.edu/publications/1947/1}} of size $128\times128\times220$ with its band 1 to band 32 are included. The hyperparameters of our method for real experimrnts are $\alpha_1 = 0.1,\alpha_2 = 0.1$, and $\alpha_3=0.01$.\par
All the experiments are conducted on the platform of Windows 10 with Intel Core i5-9400f CPU, Nvidia RTX 2080 GPU, and 16 GB RAM. 
\subsection{Simulation Experiments}
The numerical results on HSIs are illustrated in Table \ref{HSI_tab}. We can discover that S2DIP* outperforms competing methods in terms of PSNR. S2DIP is also competitive with PSNR a little lower than S2DIP*, which shows that S2DIP has convergence property to stably handle the mixed noise removal under unsupervised conditions, which avoids the semi-convergence of DIP.\par
It is notable that the SAM values of the proposed method are also superior to competing methods. This mainly attributes to the spatial-spectral constraint of the proposed framework, where high-quality spatial-spectral correlation and spectral fidelity are ensured. \par
In Fig. \ref{HSI_fig}, the denoising results on HSIs for {\bf Case 1} are displayed. Since {\bf Case 1} only contains Gaussian noise, LRTDTV, HSID-CNN, and S2DIP can both remove the noise well and obtain good visual quality, while S2DIP achieves better PSNR values. We subsequently illustrate the denoising results on HSIs for {\bf Case 5} in Fig. \ref{HSI_fig_2}, as {\bf Case 5} contains the most complex noise. We can see that LRMR, which delivers the low-rankness by matrix, is hard to totally remove the mixed noise. LRTDTV considers the tensor low-rankness, thus it achieves noise removal in partial bands, but also fails to totally remove the stripes and impulses. The supervised HSID-CNN is trained on data with only Gaussian noise. Thus, it is hard to deal with strong complex simulated noise. DIP 2D and DIP 3D are hard to fit the signal part of the observation before fitting the complex noise. Thus, DIP could not well recover clean HSIs. The proposed S2DIP achieves the best performance compared with competing methods, where the mixed noise is considerably removed and the image details are well preserved. The good visual quality of the denoising results of S2DIP can attribute to the combination of the expressive power of the CNN and the hand-crafted priors, which ensures the recovery quality.
\par
Next, we display the spectral curves of the recovered results by different methods on HSIs in Fig. \ref{HSI_spec} and Fig. \ref{HSI_spec_2}. The less oscillating spectral curves of the denoising results by S2DIP verify that the spectral fidelity is well-preserved, which outperforms competing methods. The integration of deep prior brought by the CNN and the spatial-spectral constraint by the SSTV regularization contributes to this phenomenon.
\begin{figure*}[t]
\scriptsize
\setlength{\tabcolsep}{0.9pt}
\begin{center}
\begin{tabular}{ccccccc}
\includegraphics[width=0.14\textwidth]{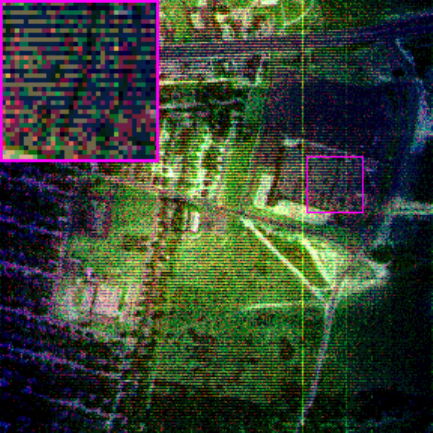}&
\includegraphics[width=0.14\textwidth]{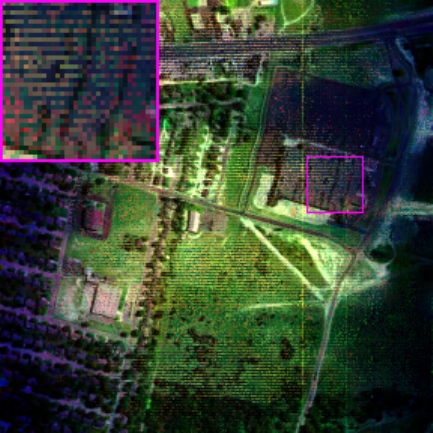}&
\includegraphics[width=0.14\textwidth]{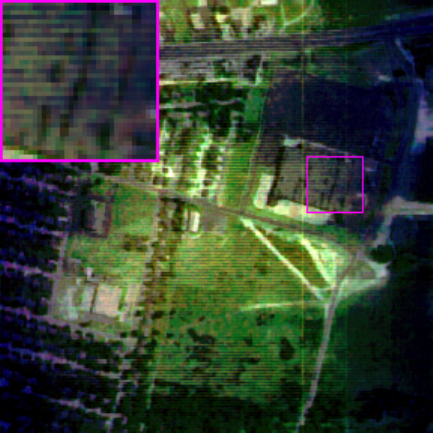}&
\includegraphics[width=0.14\textwidth]{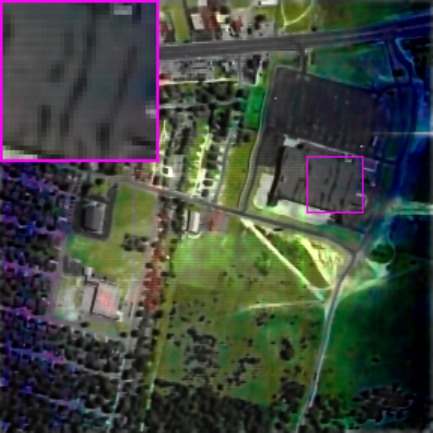}&
\includegraphics[width=0.14\textwidth]{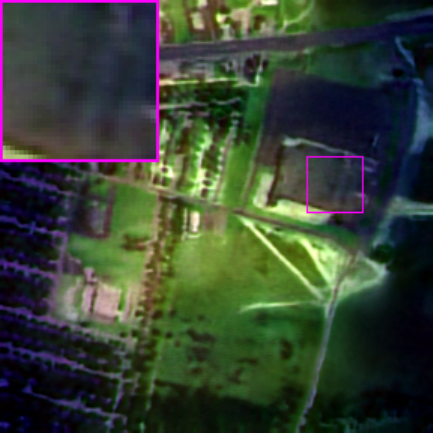}&
\includegraphics[width=0.14\textwidth]{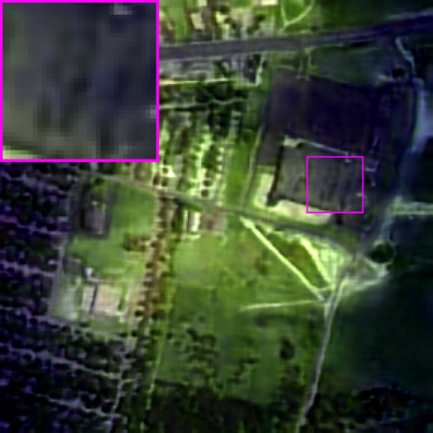}&
\includegraphics[width=0.14\textwidth]{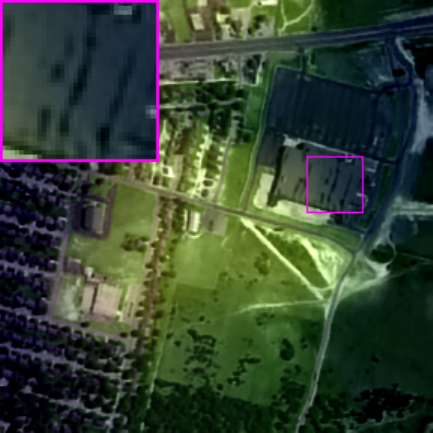}\\
\vspace{-0.3cm}
Noisy & LRMR\cite{DeepHS} & LRTDTV\cite{LRTDTV}&HSID-CNN\cite{HSID-CNN}&DIP 2D\cite{DeepHS}& DIP 3D\cite{DeepHS}&S2DIP\\
\end{tabular}
\end{center}
\caption{The recovered results by different methods for real-world noisy HSI {\it Urban} (The pseudo images consisted of the 139-th, 150-th, and 151-th bands). \label{urban}}
\end{figure*}
\begin{figure*}[!h]
\scriptsize
\setlength{\tabcolsep}{0.9pt}
\begin{center}
\begin{tabular}{ccccccc}
\includegraphics[width=0.14\textwidth]{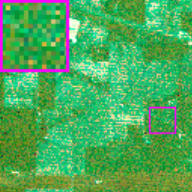}&
\includegraphics[width=0.14\textwidth]{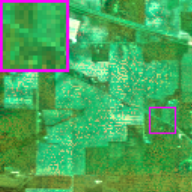}&
\includegraphics[width=0.14\textwidth]{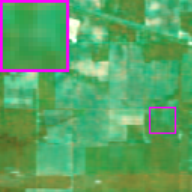}&
\includegraphics[width=0.14\textwidth]{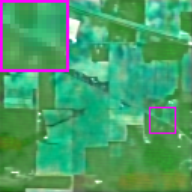}&
\includegraphics[width=0.14\textwidth]{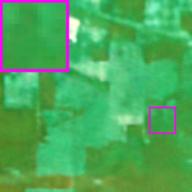}&
\includegraphics[width=0.14\textwidth]{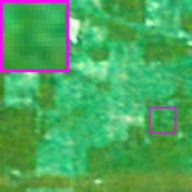}&
\includegraphics[width=0.14\textwidth]{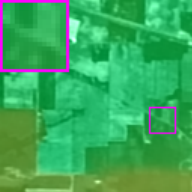}\\
\vspace{-0.3cm}
Noisy & LRMR\cite{DeepHS} & LRTDTV\cite{LRTDTV}&HSID-CNN\cite{HSID-CNN}&DIP 2D\cite{DeepHS}& DIP 3D\cite{DeepHS}&S2DIP\\
\end{tabular}
\end{center}
\caption{The recovered results by different methods for real-world noisy HSI {\it Indian} (The pseudo images consisted of the 1-st, 2-nd, and 3-rd bands). \label{indian}}
\vspace{-0.2cm}
\end{figure*}
\par
Then, we illustrate the denoising quantitative results by different methods on MSIs in Table \ref{MSI_tab}. We can see that S2DIP and S2DIP* considerably outperforms competing methods, which verify the generalization ability and effectiveness of the proposed method for various data. \par
The denoising results by different methods on MSIs for {\bf Case 5} are displayed in Fig. \ref{MSI_fig}. We can observe that S2DIP recovers the MSI and removes the complex noise well. LRMR and LRTDTV could not totally remove the noise. The HSID-CNN is trained on Gaussian noisy HSIs and thus it is hard to deal with such mixed noise in MSIs. As compared, S2DIP has a better generalization ability for mixed noise in various types of data. DIP methods remove the noise but miss some image details and edges. Note that the results of DIP are selected by referring to the GT images on the highest PSNR values. In contrast, the results of S2DIP are automatically selected based on the stopping criterion. This verifies that the proposed S2DIP favorably addresses the semi-convergence of DIP and largely enhances the denoising ability of DIP.\par
Finally, we plot the spectral curves of the recovered MSIs by different methods in Fig. \ref{MSI_spec} and Fig. \ref{MSI_spec_2}. We can see that S2DIP better preserves the nonlinear spectral curves. This could attribute to the integration of hand-crafted priors and the DIP, where the nonlinear modeling ability of CNN and the spatial-spectral constraint are well combined.
\subsection{Real Experiments}
The denoising results for real-world noisy HSI {\it Urban} are displayed in Fig. \ref{urban}. We can see that S2DIP shows better performance for removing the complex noise. Specifically, S2DIP removes the noise and preserves the image details well. LRMR and LRTDTV could not totally remove the mixed noise. DIP has over-smooth results. HSID-CNN achieves considerable results. But if one looks closely, the result of HSID-CNN remain some stripe noise.  
\par
The denoising results for {\it Indian} are illustrated in Fig. \ref{indian}. We can see that LRTDTV and DIP 2D both remove the strong noise but miss some image details. HSID-CNN removes the complex noise and achieves good results. S2DIP also successfully and reasonably restores the image structure, where the complex noise is removed and the image details and colors are well preserved. 
\par
The good performance of S2DIP for real noise removal is mainly due to the combination of deep prior and hand-crafted priors. The natural image structure can be well preserved by the DIP and the mixed noise can be fully removed by the spatial-spectral constraint and the robust noise modeling. 
\par
We can find that in real experiments, the performance of HSID-CNN is much better than that of simulation experiments. This is because that the number of noisy bands in real noisy HSIs is relatively small. HSID-CNN can explore the spatial-spectral information from other clean bands for denoising. The proposed unsupervised S2DIP could achieve even better results than the supervised HSID-CNN for real noise removal. Also, the denoising results by S2DIP can be selected by setting different model parameters. Supervised methods like HSID-CNN use one well-trained CNN for denoising and thus could only have one chance for denoising. In other words, S2DIP may be relatively more flexible in real applications. \par
\begin{figure}[!t]
\scriptsize
\setlength{\tabcolsep}{0.9pt}
\begin{center}
\begin{tabular}{cccc}
\includegraphics[width=0.24\linewidth]{om1case5.pdf}&
 \includegraphics[width=0.24\linewidth]{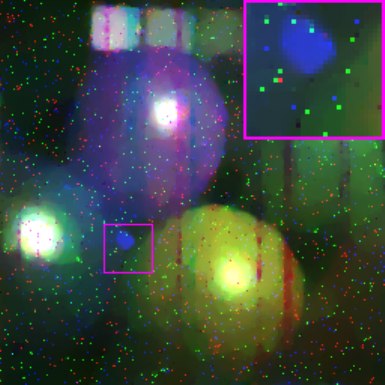}&
\includegraphics[width=0.24\linewidth]{om1case5S2DIP.pdf}&
\includegraphics[width=0.24\linewidth]{om1gtcase5.pdf}\\
\vspace{0.05cm}
PSNR 13.200 dB&PSNR 22.187 dB&PSNR 37.367 dB&PSNR Inf\\
\includegraphics[width=0.24\linewidth]{om9case5.pdf}&
\includegraphics[width=0.24\linewidth]{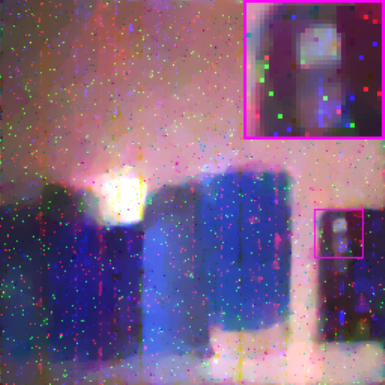}&
\includegraphics[width=0.24\linewidth]{om9case5S2DIP.pdf}&
\includegraphics[width=0.24\linewidth]{om9gtcase5.pdf}\\
PSNR 13.934 dB&PSNR 22.625 dB&PSNR 37.834 dB&PSNR Inf\\
\vspace{-0.2cm}
Noisy&HSSTV \cite{HSSTV}&S2DIP&GT\\
\end{tabular}
\end{center}
\caption{The denoising MSIs {\it balloons} and {\it Cups} consisted of the 1-st, 10-th, and 30-th bands by different methods.\label{hsstv_fig}}
\end{figure}
\begin{table}[t]
\caption{The quantitative results by different methods. S2DIP wo. sparsity denotes the proposed method without the sparse term. S2DIP wo. SSTV denotes the proposed method without the TV and SSTV term.}\vspace{-0.4cm}
\begin{center}
\scriptsize
\setlength{\tabcolsep}{5pt}
\begin{spacing}{1.00}
\begin{tabular}{ccccccc}
\toprule
\multirow{3}*{Method}&\multicolumn{3}{c}{\it Balloons {\bf Case 5}}&\multicolumn{3}{c}{\it Cups {\bf Case 5}}\\
\cmidrule{2-7}
~&PSNR&SSIM&SAM&\;\;PSNR&SSIM&SAM\\
\midrule
S2DIP wo. sparsity & 26.116 & 0.816 & 0.258 &\;\; 28.952 & 0.955 & 0.076 \\
S2DIP wo. SSTV & 34.775 & 0.949 & 0.167&\;\; 33.729 & 0.958 & 0.059 \\
S2DIP& \bf{37.367} & \bf{0.971} & \bf{0.125} &\;\; \bf{37.834} & \bf{0.988} & \bf{0.034} \\
\bottomrule
\end{tabular}
\end{spacing}
\end{center}
\label{tab_l1}\vspace{-0.4cm}
\end{table}
\section{Discussions} \label{sec:Dis}
\subsection{Effectiveness of Deep Prior}
Our method simultaneously uses the unsupervised deep prior, the spatial-spectral image prior, and the sparse noise prior. In this section, we illustrate the effectiveness of the deep prior. Specifically, we remove the deep prior of CNN in the proposed model (\ref{main_model}) to clarify its influence. Here, the underlying clean HSI could be updated via ADMM. Note that the proposed method degenerates to the hybrid spatial-spectral total variation (HSSTV) \cite{HSSTV} method without the deep prior. The HSSTV considers the TV and SSTV on the clean HSI and remove the sparse noise via minimizing its $\ell_1$-norm. The difference between HSSTV and the proposed method is that we additionally use the DIP while HSSTV just employs TV and SSTV regularizers in a traditional optimization model.
\par
We conduct the comparison between HSSTV and S2DIP on {\it Balloons} and {\it Cups} for {\bf Case 5}. Fig. \ref{hsstv_fig} shows the denoising results. We can see that S2DIP removes the complex noise and preserves the image details and edges well. In contrast, HSSTV could not completely remove the impulse noise and stripes, while the image details are also missed. This verifies the effectiveness of the deep prior, which can capture the natural image structure.  S2DIP faithfully combines the deep prior of CNN and the hand-crafted prior, taking both advantages of the high representation ability of CNN and the spatial-spectral constraint delivered by the SSTV in an unsupervised manner.
\subsection{Effectiveness of Hand-Crafted Priors}
To address the semi-convergence of DIP and improve its denoising performance on HSI, we suggest two hand-crafted priors in DIP, i.e., the SSTV term and the sparse term. This section verifies their effectiveness. Specifically, we compare the proposed method with and without the SSTV term and the sparse term to clarify their influence. The results are illustrated in Table \ref{tab_l1}. We can see that S2DIP considerably outperforms S2DIP wo. SSSTV and S2DIP wo. sparsity, which verifies the effectiveness of the hand-crafted priors in the DIP framework. The deep prior, the spatial-spectral local smooth prior, and the sparse noise prior are organically combined to handle the mixed noise removal in HSI. Notably, S2DIP wo. sparsity has relatively weak performance, which shows that the sparse modeling of the complex noise is very essential. In fact, the HSI noise in {\bf Case 5} contains many sparse outliers, which can not be fully removed if only considering Gaussian noise. Thus, introducing the sparse term can largely improve the robustness and effectiveness of the unsupervised DIP for HSI mixed noise removal.
\begin{figure*}[!h]
\scriptsize
\setlength{\tabcolsep}{0.9pt}
\begin{center}
\begin{tabular}{ccccccc}
\includegraphics[width=0.14\textwidth]{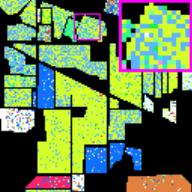}&
\includegraphics[width=0.14\textwidth]{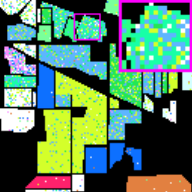}&
\includegraphics[width=0.14\textwidth]{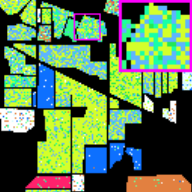}&
\includegraphics[width=0.14\textwidth]{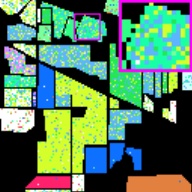}&
\includegraphics[width=0.14\textwidth]{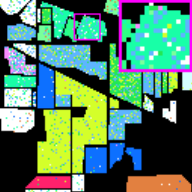}&
\includegraphics[width=0.14\textwidth]{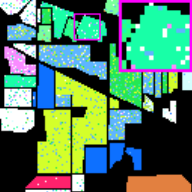}&
\includegraphics[width=0.14\textwidth]{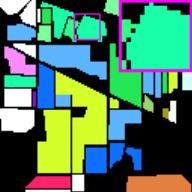}\\
\vspace{0.05cm}
Accuracy 0.757&Accuracy 0.895&Accuracy 0.770&Accuracy 0.881&Accuracy {0.907}&Accuracy {\bf 0.935}&Accuracy 1.000\\
\vspace{-0.3cm}
Noisy& LRMR\cite{DeepHS} & LRTDTV\cite{LRTDTV}&HSID-CNN \cite{HSID-CNN}&DIP 2D\cite{DeepHS}&S2DIP&GT\\
\end{tabular}
\end{center}
\caption{The HSI classification results using SVM on the denoising results by different denoising methods. Different colors refer to different classifications. The {\bf best} accuracy rate is highlighted by {\bf boldface}.\label{class_fig}}\vspace{-0.2cm}
\end{figure*}
\begin{figure}[t]
\scriptsize
\setlength{\tabcolsep}{0.9pt}
\begin{center}
\begin{tabular}{c}
\vspace{-0.3cm}
\includegraphics[width=1\linewidth]{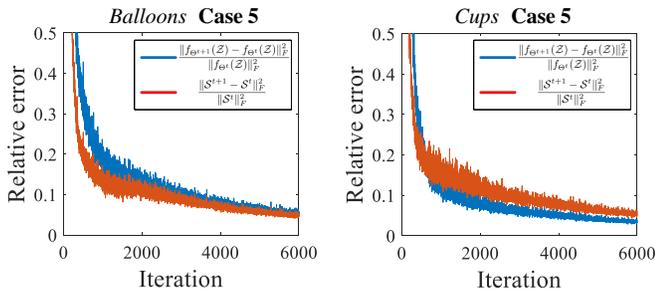}
\end{tabular}
\end{center}
\caption{The relative error of variables with respect to iteration.\label{convergence}}
\end{figure}
\begin{figure}[!h]
\scriptsize
\setlength{\tabcolsep}{0.9pt}
\begin{center}
\begin{tabular}{cccc}
\includegraphics[width=0.24\linewidth]{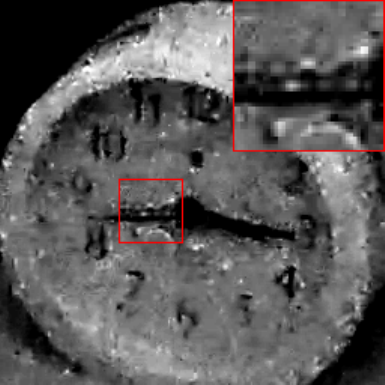}&
\includegraphics[width=0.24\linewidth]{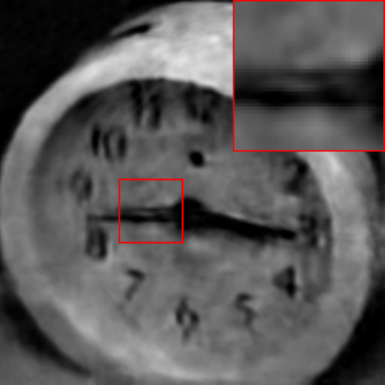}&
\includegraphics[width=0.24\linewidth]{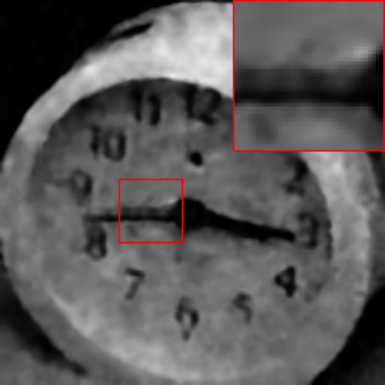}&
\includegraphics[width=0.24\linewidth]{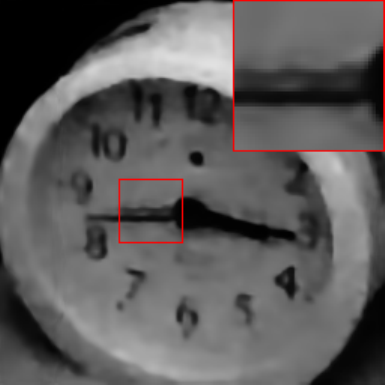}\\
\includegraphics[width=0.24\linewidth]{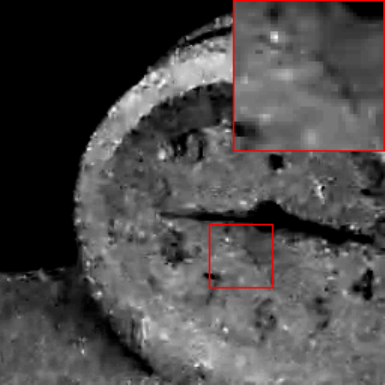}&
\includegraphics[width=0.24\linewidth]{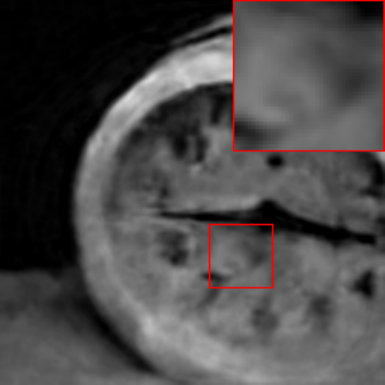}&
\includegraphics[width=0.24\linewidth]{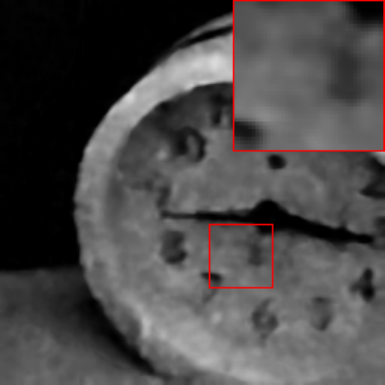}&
\includegraphics[width=0.24\linewidth]{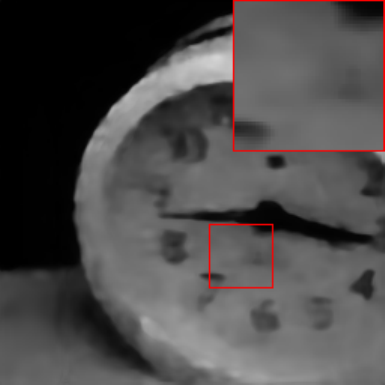}\\
\vspace{-0.2cm}
Noisy & DIP 2D\cite{DeepHS} & DIP 3D\cite{DeepHS} & S2DIP\\
\end{tabular}
\end{center}
\caption{Two denoising video frames by different methods.\label{vid}}
\vspace{-0.3cm}
\end{figure}
\subsection{Influence on Subsequent Applications}
This section verifies that using S2DIP for HSI denoising can improve the performance of subsequent applications. We consider the HSI classification \cite{Meng_2} using the support vector machine (SVM). We use the denoising results of the real-world noisy HSI {\it Indian} by different methods to conduct the HSI classification. $10\%$ of the samples are set as training set while others are testing set. The results are shown in Fig. \ref{class_fig}. We can see that the classification result on the denoising HSI by S2DIP has the best visual quality and accuracy, which verifies the effectiveness and superiority of S2DIP over competing denoising methods.
\subsection{Convergence Analysis}
To test the convergence of the ADMM Algorithm \ref{alg}, we plot the relative error of variables with respect to the iteration number in Fig. \ref{convergence}. The downward trend of the curves verifies the convergence behavior of our method, which favorably addresses the semi-convergence of DIP. This can attribute to the SSTV and the sparse term which respectively considers the HSI spatial-spectral local smoothness and the sparsity of the structural noise. Thus, the proposed model can accurately separate the noise and the clean HSI, resulting in stable convergence.
\subsection{The Generalization of S2DIP}
In this subsection, we discuss the generalization of the proposed method. Here, we consider video data\footnote{The real noisy video is captured by our personal device under extremely low light conditions.} which also contains three-dimensional information. In low light conditions, the videos will be inevitably corrupted by dynamic noise \cite{Wang_ICCV}, see Fig. \ref{vid}.
\par
Due to the complexity and irregularity of the real noise, video denoising is extremely challenging. For the heavily corrupted video frames, we use DIP and our method for its restoration. Fig. \ref{vid} displays the denoising results, where S2DIP reasonably recovers the clean video and preserves the temporal fidelity. The results of DIP methods, which are manually selected due to the semi-convergence, remain notable noise. The above results illustrate the generalization of S2DIP to handle different types of data and various noise. 
\section{Conclusion} \label{sec:Con}
In this paper, we propose the S2DIP for hyperspectral mixed noise removal. The SSTV regularization is considered to fully explore the HSI spatial-spectral local smooth prior to improve the effectiveness of the unsupervised DIP. The $\ell_1$-norm sparse term is considered to remove the sparse noise. To address the proposed model, we develop an ADMM algorithm. Experimental results on synthetic data and real data demonstrate the effectiveness of the proposed method, which outperforms state-of-the-art methods. Meanwhile, the proposed method favorably alleviates the semi-convergence problem, which always exists in the DIP framework. Our future work would contain the usage of unsupervised CNN and its combination with model-based methods for more applications, especially for multi-dimensional image processing.
{
\bibliographystyle{ieeetran}
\bibliography{refference}

\begin{thebibliography}{10}
\providecommand{\url}[1]{#1}
\csname url@samestyle\endcsname
\providecommand{\newblock}{\relax}
\providecommand{\bibinfo}[2]{#2}
\providecommand{\BIBentrySTDinterwordspacing}{\spaceskip=0pt\relax}
\providecommand{\BIBentryALTinterwordstretchfactor}{4}
\providecommand{\BIBentryALTinterwordspacing}{\spaceskip=\fontdimen2\font plus
\BIBentryALTinterwordstretchfactor\fontdimen3\font minus
  \fontdimen4\font\relax}
\providecommand{\BIBforeignlanguage}[2]{{%
\expandafter\ifx\csname l@#1\endcsname\relax
\typeout{** WARNING: IEEEtran.bst: No hyphenation pattern has been}%
\typeout{** loaded for the language `#1'. Using the pattern for}%
\typeout{** the default language instead.}%
\else
\language=\csname l@#1\endcsname
\fi
#2}}
\providecommand{\BIBdecl}{\relax}
\BIBdecl

\bibitem{DeepHS}
O.~{Sidorov} and J.~Y. {Hardeberg}, ``Deep hyperspectral prior: Single-image
  denoising, inpainting, super-resolution,'' in \emph{2019 IEEE/CVF
  International Conference on Computer Vision Workshop (ICCVW)}, 2019, pp.
  3844--3851.

\bibitem{Wang2016}
G.~Wang, Y.~Zhang, B.~He, and K.~T. Chong, ``A framework of target detection in
  hyperspectral imagery based on blind source extraction,'' \emph{IEEE Journal
  of Selected Topics in Applied Earth Observations and Remote Sensing}, vol.~9,
  no.~2, pp. 835--844, 2016.

\bibitem{Liu2017Target}
Y.~Liu, G.~Gao, and Y.~Gu, ``Tensor matched subspace detector for hyperspectral
  target detection,'' \emph{IEEE Transactions on Geoscience and Remote
  Sensing}, vol.~55, no.~4, pp. 1967--1974, 2017.

\bibitem{Jia2018}
S.~Jia, L.~Shen, J.~Zhu, and Q.~Li, ``A 3-{D} gabor phase-based coding and
  matching framework for hyperspectral imagery classification,'' \emph{IEEE
  Transactions on Cybernetics}, vol.~48, no.~4, pp. 1176--1188, 2018.

\bibitem{TNNLS_class1}
N.~{Akhtar} and A.~{Mian}, ``Nonparametric coupled bayesian dictionary and
  classifier learning for hyperspectral classification,'' \emph{IEEE
  Transactions on Neural Networks and Learning Systems}, vol.~29, no.~9, pp.
  4038--4050, 2018.

\bibitem{Meng_2}
X.~Cao, J.~Yao, Z.~Xu, and D.~Meng, ``Hyperspectral image classification with
  convolutional neural network and active learning,'' \emph{IEEE Transactions
  on Geoscience and Remote Sensing}, vol.~58, no.~7, pp. 4604--4616, 2020.

\bibitem{Wang_2}
Y.~Wang, L.~Lin, Q.~Zhao, T.~Yue, D.~Meng, and Y.~Leung, ``Compressive sensing
  of hyperspectral images via joint tensor tucker decomposition and weighted
  total variation regularization,'' \emph{IEEE Geoscience and Remote Sensing
  Letters}, vol.~14, no.~12, pp. 2457--2461, 2017.

\bibitem{He_2}
W.~He, H.~Zhang, and L.~Zhang, ``Total variation regularized reweighted sparse
  nonnegative matrix factorization for hyperspectral unmixing,'' \emph{IEEE
  Transactions on Geoscience and Remote Sensing}, vol.~55, no.~7, pp.
  3909--3921, 2017.

\bibitem{Meng_3}
J.~Yao, D.~Hong, L.~Xu, D.~Meng, J.~Chanussot, and Z.~Xu, ``Sparsity-enhanced
  convolutional decomposition: A novel tensor-based paradigm for blind
  hyperspectral unmixing,'' \emph{IEEE Transactions on Geoscience and Remote
  Sensing}, pp. 1--14, 2021.

\bibitem{Wang_1}
K.~Wang, Y.~Wang, X.-L. Zhao, J.~C.-W. Chan, Z.~Xu, and D.~Meng,
  ``Hyperspectral and multispectral image fusion via nonlocal low-rank tensor
  decomposition and spectral unmixing,'' \emph{IEEE Transactions on Geoscience
  and Remote Sensing}, vol.~58, no.~11, pp. 7654--7671, 2020.

\bibitem{ASSTV}
W.~{He}, H.~{Zhang}, H.~{Shen}, and L.~{Zhang}, ``Hyperspectral image denoising
  using local low-rank matrix recovery and global spatial–spectral total
  variation,'' \emph{IEEE Journal of Selected Topics in Applied Earth
  Observations and Remote Sensing}, vol.~11, no.~3, pp. 713--729, 2018.

\bibitem{SSTV}
H.~K. {Aggarwal} and A.~{Majumdar}, ``Hyperspectral image denoising using
  spatio-spectral total variation,'' \emph{IEEE Geoscience and Remote Sensing
  Letters}, vol.~13, no.~3, pp. 442--446, 2016.

\bibitem{HSSTV}
S.~{Takeyama}, S.~{Ono}, and I.~{Kumazawa}, ``Mixed noise removal for
  hyperspectral images using hybrid spatio-spectral total variation,'' in
  \emph{2019 IEEE International Conference on Image Processing (ICIP)}, 2019,
  pp. 3128--3132.

\bibitem{DIP-tv}
J.~{Liu}, Y.~{Sun}, X.~{Xu}, and U.~S. {Kamilov}, ``Image restoration using
  total variation regularized deep image prior,'' in \emph{ICASSP 2019 - 2019
  IEEE International Conference on Acoustics, Speech and Signal Processing
  (ICASSP)}, 2019, pp. 7715--7719.

\bibitem{sparse1}
J.~{Mairal}, F.~{Bach}, J.~{Ponce}, G.~{Sapiro}, and A.~{Zisserman},
  ``Non-local sparse models for image restoration,'' in \emph{2009 IEEE 12th
  International Conference on Computer Vision}, 2009, pp. 2272--2279.

\bibitem{sparse2}
W.~{Dong}, L.~{Zhang}, G.~{Shi}, and X.~{Li}, ``Nonlocally centralized sparse
  representation for image restoration,'' \emph{IEEE Transactions on Image
  Processing}, vol.~22, no.~4, pp. 1620--1630, 2013.

\bibitem{BM4D}
M.~{Maggioni}, V.~{Katkovnik}, K.~{Egiazarian}, and A.~{Foi}, ``Nonlocal
  transform-domain filter for volumetric data denoising and reconstruction,''
  \emph{IEEE Transactions on Image Processing}, vol.~22, no.~1, pp. 119--133,
  2013.

\bibitem{LRMR}
H.~{Zhang}, W.~{He}, L.~{Zhang}, H.~{Shen}, and Q.~{Yuan}, ``Hyperspectral
  image restoration using low-rank matrix recovery,'' \emph{IEEE Transactions
  on Geoscience and Remote Sensing}, vol.~52, no.~8, pp. 4729--4743, 2014.

\bibitem{tvLRMR}
W.~{He}, H.~{Zhang}, L.~{Zhang}, and H.~{Shen}, ``Total-variation-regularized
  low-rank matrix factorization for hyperspectral image restoration,''
  \emph{IEEE Transactions on Geoscience and Remote Sensing}, vol.~54, no.~1,
  pp. 178--188, 2016.

\bibitem{LRTDTV}
Y.~{Wang}, J.~{Peng}, Q.~{Zhao}, Y.~{Leung}, X.~{Zhao}, and D.~{Meng},
  ``Hyperspectral image restoration via total variation regularized low-rank
  tensor decomposition,'' \emph{IEEE Journal of Selected Topics in Applied
  Earth Observations and Remote Sensing}, vol.~11, no.~4, pp. 1227--1243, 2018.

\bibitem{ChangYi2017}
Y.~{Chang}, L.~{Yan}, and S.~{Zhong}, ``Hyper-laplacian regularized
  unidirectional low-rank tensor recovery for multispectral image denoising,''
  in \emph{2017 IEEE Conference on Computer Vision and Pattern Recognition
  (CVPR)}, 2017, pp. 5901--5909.

\bibitem{Low_rank_JSTSP}
A.~{Karami}, M.~{Yazdi}, and A.~{Zolghadre Asli}, ``Noise reduction of
  hyperspectral images using kernel non-negative tucker decomposition,''
  \emph{IEEE Journal of Selected Topics in Signal Processing}, vol.~5, no.~3,
  pp. 487--493, 2011.

\bibitem{XieQi_PAMI}
Q.~{Xie}, Q.~{Zhao}, D.~{Meng}, and Z.~{Xu}, ``Kronecker-basis-representation
  based tensor sparsity and its applications to tensor recovery,'' \emph{IEEE
  Transactions on Pattern Analysis and Machine Intelligence}, vol.~40, no.~8,
  pp. 1888--1902, 2018.

\bibitem{TNNLS_Meng}
X.~{Chen}, Z.~{Han}, Y.~{Wang}, Q.~{Zhao}, D.~{Meng}, L.~{Lin}, and Y.~{Tang},
  ``A generalized model for robust tensor factorization with noise modeling by
  mixture of gaussians,'' \emph{IEEE Transactions on Neural Networks and
  Learning Systems}, vol.~29, no.~11, pp. 5380--5393, 2018.

\bibitem{Dic_HSI}
X.~Gong, W.~Chen, and J.~Chen, ``A low-rank tensor dictionary learning method
  for hyperspectral image denoising,'' \emph{IEEE Transactions on Signal
  Processing}, vol.~68, pp. 1168--1180, 2020.

\bibitem{CVPR_Dic}
Y.~{Peng}, D.~{Meng}, Z.~{Xu}, C.~{Gao}, Y.~{Yang}, and B.~{Zhang},
  ``Decomposable nonlocal tensor dictionary learning for multispectral image
  denoising,'' in \emph{2014 IEEE Conference on Computer Vision and Pattern
  Recognition}, 2014, pp. 2949--2956.

\bibitem{LLRGTV}
W.~He, H.~Zhang, H.~Shen, and L.~Zhang, ``Hyperspectral image denoising using
  local low-rank matrix recovery and global spatial–spectral total
  variation,'' \emph{IEEE Journal of Selected Topics in Applied Earth
  Observations and Remote Sensing}, vol.~11, no.~3, pp. 713--729, 2018.

\bibitem{Lina_1}
L.~Zhuang, X.~Fu, M.~K. Ng, and J.~M. Bioucas-Dias, ``Hyperspectral image
  denoising based on global and nonlocal low-rank factorizations,'' \emph{IEEE
  Transactions on Geoscience and Remote Sensing}, pp. 1--17, 2021.

\bibitem{Lina_2}
L.~Zhuang, L.~Gao, B.~Zhang, X.~Fu, and J.~M. Bioucas-Dias, ``Hyperspectral
  image denoising and anomaly detection based on low-rank and sparse
  representations,'' \emph{IEEE Transactions on Geoscience and Remote Sensing},
  pp. 1--17, 2020.

\bibitem{He_3}
W.~He, H.~Zhang, H.~Shen, and L.~Zhang, ``Hyperspectral image denoising using
  local low-rank matrix recovery and global spatial–spectral total
  variation,'' \emph{IEEE Journal of Selected Topics in Applied Earth
  Observations and Remote Sensing}, vol.~11, no.~3, pp. 713--729, 2018.

\bibitem{He_4}
W.~He, H.~Zhang, L.~Zhang, and H.~Shen, ``Total-variation-regularized low-rank
  matrix factorization for hyperspectral image restoration,'' \emph{IEEE
  Transactions on Geoscience and Remote Sensing}, vol.~54, no.~1, pp. 178--188,
  2016.

\bibitem{TIP_2017}
K.~H. {Jin}, M.~T. {McCann}, E.~{Froustey}, and M.~{Unser}, ``Deep
  convolutional neural network for inverse problems in imaging,'' \emph{IEEE
  Transactions on Image Processing}, vol.~26, no.~9, pp. 4509--4522, 2017.

\bibitem{NIPS2016}
X.~Mao, C.~Shen, and Y.-B. Yang, ``Image restoration using very deep
  convolutional encoder-decoder networks with symmetric skip connections,'' in
  \emph{Advances in Neural Information Processing Systems 29}, 2016, pp.
  2802--2810.

\bibitem{ResNet}
K.~{He}, X.~{Zhang}, S.~{Ren}, and J.~{Sun}, ``Deep residual learning for image
  recognition,'' in \emph{2016 IEEE Conference on Computer Vision and Pattern
  Recognition (CVPR)}, 2016, pp. 770--778.

\bibitem{Deep_HSI_sharpening}
R.~{Dian}, S.~{Li}, A.~{Guo}, and L.~{Fang}, ``Deep hyperspectral image
  sharpening,'' \emph{IEEE Transactions on Neural Networks and Learning
  Systems}, vol.~29, no.~11, pp. 5345--5355, 2018.

\bibitem{TNNLS_2014}
P.~{Zhong} and R.~{Wang}, ``Jointly learning the hybrid {CRF} and {MLR} model
  for simultaneous denoising and classification of hyperspectral imagery,''
  \emph{IEEE Transactions on Neural Networks and Learning Systems}, vol.~25,
  no.~7, pp. 1319--1334, 2014.

\bibitem{2019Dong}
W.~{Dong}, H.~{Wang}, F.~{Wu}, G.~{Shi}, and X.~{Li}, ``Deep spatial–spectral
  representation learning for hyperspectral image denoising,'' \emph{IEEE
  Transactions on Computational Imaging}, vol.~5, no.~4, pp. 635--648, 2019.

\bibitem{HSI-DeNet}
Y.~{Chang}, L.~{Yan}, H.~{Fang}, S.~{Zhong}, and W.~{Liao}, ``H{SI}-{D}e{N}et:
  Hyperspectral image restoration via convolutional neural network,''
  \emph{IEEE Transactions on Geoscience and Remote Sensing}, vol.~57, no.~2,
  pp. 667--682, 2019.

\bibitem{HSID-CNN}
Q.~{Yuan}, Q.~{Zhang}, J.~{Li}, H.~{Shen}, and L.~{Zhang}, ``Hyperspectral
  image denoising employing a spatial–spectral deep residual convolutional
  neural network,'' \emph{IEEE Transactions on Geoscience and Remote Sensing},
  vol.~57, no.~2, pp. 1205--1218, 2019.

\bibitem{TowardUni}
Y.~Chang, M.~Chen, L.~Yan, X.-L. Zhao, Y.~Li, and S.~Zhong, ``Toward universal
  stripe removal via wavelet-based deep convolutional neural network,''
  \emph{IEEE Transactions on Geoscience and Remote Sensing}, vol.~PP, pp.
  1--18, 12 2019.

\bibitem{TNNLS_2020_HSI}
K.~{Wei}, Y.~{Fu}, and H.~{Huang}, ``3-{D} quasi-recurrent neural network for
  hyperspectral image denoising,'' \emph{IEEE Transactions on Neural Networks
  and Learning Systems}, pp. 1--13, 2020.

\bibitem{DIP}
V.~{Lempitsky}, A.~{Vedaldi}, and D.~{Ulyanov}, ``Deep image prior,'' in
  \emph{2018 IEEE/CVF Conference on Computer Vision and Pattern Recognition},
  2018, pp. 9446--9454.

\bibitem{Yuan_2}
Q.~Yuan, L.~Zhang, and H.~Shen, ``Hyperspectral image denoising employing a
  spectral–spatial adaptive total variation model,'' \emph{IEEE Transactions
  on Geoscience and Remote Sensing}, vol.~50, no.~10, pp. 3660--3677, 2012.

\bibitem{He_1}
W.~He, N.~Yokoya, L.~Yuan, and Q.~Zhao, ``Remote sensing image reconstruction
  using tensor ring completion and total variation,'' \emph{IEEE Transactions
  on Geoscience and Remote Sensing}, vol.~57, no.~11, pp. 8998--9009, 2019.

\bibitem{8894531}
H.~Zhang, L.~Liu, W.~He, and L.~Zhang, ``Hyperspectral image denoising with
  total variation regularization and nonlocal low-rank tensor decomposition,''
  \emph{IEEE Transactions on Geoscience and Remote Sensing}, vol.~58, no.~5,
  pp. 3071--3084, 2020.

\bibitem{Zhang_1}
H.~Zhang, J.~Cai, W.~He, H.~Shen, and L.~Zhang, ``Double low-rank matrix
  decomposition for hyperspectral image denoising and destriping,'' \emph{IEEE
  Transactions on Geoscience and Remote Sensing}, pp. 1--19, 2021.

\bibitem{Yuan_1}
Q.~Zhang, Q.~Yuan, J.~Li, X.~Liu, H.~Shen, and L.~Zhang, ``Hybrid noise removal
  in hyperspectral imagery with a spatial–spectral gradient network,''
  \emph{IEEE Transactions on Geoscience and Remote Sensing}, vol.~57, no.~10,
  pp. 7317--7329, 2019.

\bibitem{Meng_1}
X.~Cao, X.~Fu, C.~Xu, and D.~Meng, ``Deep spatial-spectral global reasoning
  network for hyperspectral image denoising,'' \emph{IEEE Transactions on
  Geoscience and Remote Sensing}, pp. 1--14, 2021.

\bibitem{P3D}
Z.~{Qiu}, T.~{Yao}, and T.~{Mei}, ``Learning spatio-temporal representation
  with pseudo-3d residual networks,'' in \emph{2017 IEEE International
  Conference on Computer Vision (ICCV)}, 2017, pp. 5534--5542.

\bibitem{ADAM}
D.~Kingma and J.~Ba, ``{ADAM}: A method for stochastic optimization,''
  \emph{International Conference on Learning Representations}, 12 2014.

\bibitem{cave}
F.~{Yasuma}, T.~{Mitsunaga}, D.~{Iso}, and S.~K. {Nayar}, ``Generalized
  assorted pixel camera: Postcapture control of resolution, dynamic range, and
  spectrum,'' \emph{IEEE Transactions on Image Processing}, vol.~19, no.~9, pp.
  2241--2253, 2010.

\bibitem{sam}
B.~R. {Shivakumar} and S.~V. {Rajashekararadhya}, ``Performance évaluation of
  spectral angle mapper and spectral correlation mapper classifiers over
  multiple remote sensor data,'' in \emph{2017 Second International Conference
  on Electrical, Computer and Communication Technologies (ICECCT)}, 2017, pp.
  1--6.

\bibitem{Wang_ICCV}
W.~{Wang}, X.~{Chen}, C.~{Yang}, X.~{Li}, X.~{Hu}, and T.~{Yue}, ``Enhancing
  low light videos by exploring high sensitivity camera noise,'' in \emph{2019
  IEEE/CVF International Conference on Computer Vision (ICCV)}, 2019, pp.
  4110--4118.

\end{thebibliography}
}
\end{document}